\documentclass[12pt]{article}
\usepackage{amsmath}
\usepackage{graphicx}
\usepackage{enumerate}
\usepackage{natbib}
\usepackage{url} 
\usepackage{amssymb}
\usepackage{ntheorem}
\usepackage{cleveref}
\usepackage{algorithm}
\usepackage{algpseudocode}
\usepackage{setspace}
\usepackage{multirow}
\usepackage{booktabs}
\usepackage{subfigure}  
\usepackage{xcolor}
\usepackage[title]{appendix}
\allowdisplaybreaks
\newcommand{\blind}{0}
\newcommand{\bd}{\boldsymbol}
\newcommand{\mb}{\mathbf}

\newtheorem{theorem}{Theorem}

\DeclareMathOperator*{\argmin}{arg\,min}

\begin{document}

\def\spacingset#1{\renewcommand{\baselinestretch}%
{#1}\small\normalsize} \spacingset{1}


\if0\blind
{
  \title{\bf High-Dimensional Dynamic Covariance Models with
Random Forests}
	\author{Shuguang Yu$^{1,\dag}$, Fan Zhou$^{1,\dag}$, Yingjie Zhang$^{2,\dag}$, Ziqi Chen$^{2*}$ and Hongtu Zhu$^{3*}$
\hspace{.2cm}\\
\\
		$^1$School of Statistics and Management,\\ Shanghai University of Finance and Economics\\
		$^2$School of Statistics, KLATASDS-MOE, East China Normal University\\ 
        $^3$Departments of Biostatistics,
Statistics, Computer Science, and Genetics,\\
The University of North Carolina at Chapel Hill\\
$^\dag$Equal contributions\\
		$^*$Correspondence: zqchen@fem.ecnu.edu.cn; htzhu@email.unc.edu
		} 
  \date{}
  \maketitle
} \fi

\bigskip
\begin{abstract}

This paper introduces a novel nonparametric method for estimating high-dimensional dynamic covariance matrices with multiple conditioning covariates, leveraging random forests and supported by robust theoretical guarantees. Unlike traditional static methods, our dynamic nonparametric covariance models effectively capture distributional heterogeneity. Furthermore, unlike kernel-smoothing methods, which are restricted to a single conditioning covariate, our approach accommodates multiple covariates in a fully nonparametric framework. 
To the best of our knowledge, this is the first method to use random forests for estimating high-dimensional dynamic covariance matrices. In high-dimensional settings, we establish uniform consistency theory, providing nonasymptotic error rates and model selection properties, even when the response dimension grows sub-exponentially with the sample size. These results hold uniformly across a range of conditioning variables.  
The method’s effectiveness is demonstrated through simulations and a stock dataset analysis, highlighting its ability to model complex dynamics in high-dimensional scenarios.


\end{abstract}

\noindent%
\vfill

\spacingset{1.9} 
\section{Introduction}
\label{sec:intro}


Let $\bd{Y} = (Y_1, \ldots, Y_p)^\top$ be a $p$-dimensional random vector, and let $\bd{U} = (U_1, \ldots, U_d)^\top$ denote the associated conditioning covariates. We model the conditional mean and covariance of $\bd{Y}$ given $\bd{U}$ nonparametrically as $\bd{m}(\bd{U}) = (m_1(\bd{U}), \dots, m_p(\bd{U}))^\top$ and $\bd{\Sigma}(\bd{U}) = \{\sigma_{jr}(\bd{U})\}_{p \times p}$, respectively, where $\sigma_{jr}(\bd{U}) = \mathrm{Cov}(Y_j, Y_r \mid \bd{U})$. 
Suppose we observe random samples $\{\bd{Y}_i, \bd{U}_i\}_{i=1}^n$ drawn from the joint distribution of $\{\bd{Y}, \bd{U}\}$. Our primary goal is to estimate the conditional covariance matrix $\bd{\Sigma}(\bd{U})$ nonparametrically, with particular interest in the high-dimensional regime where $n \ll p$. 
Previous studies \citep{Chen_2013, chen2016dynamic, wang2021nonparametric} have proposed using kernel-smoothing techniques to estimate $\bd{\Sigma}(\cdot)$ when the conditioning covariate $\bd{U}$ is one-dimensional. However, these methods become less effective when $\bd{U}$ is multivariate due to the well-known ``curse of dimensionality'', which significantly limits the performance of kernel-smoothing techniques in higher dimensions. This challenge necessitates alternative approaches to handle multivariate covariates effectively in high-dimensional settings.


High-dimensional dynamic covariance matrices, which capture complex relationships and dependencies among numerous variables, are essential in statistics, machine learning, and data science \citep{chen2016dynamic, Jiang2020, ke2022high, JMLR:v23:21-0795, jiang2024dynamic}. In practice, these matrices often vary as a function of multiple covariates. 
For instance, in functional brain connectivity, the co-occurrence of brain activity across regions is modeled as a function of covariates encoding experimental conditions or subgroups, revealing variability in brain function \citep{10.3389/fnins.2017.00696}. In finance, dependencies among stocks change with covariates like returns on Fama-French factors, reflecting dynamic market behaviors \citep{chen2019new}. 
This motivates the development of nonparametric models for high-dimensional dynamic covariance matrices, where the entries depend on multiple conditioning covariates. Nonparametric modeling mitigates the risk of model misspecification and offers the flexibility to adapt to complex data structures. 
Nevertheless, estimating such covariance matrices is challenging due to the ``curse of dimensionality" in multivariate nonparametric regression and the high-dimensional regime. To our knowledge, this paper is the first to propose a fully nonparametric framework for estimating high-dimensional dynamic covariance matrices in settings where the covariate space exceeds two or three dimensions.

In this paper, we propose leveraging random forests to estimate high-dimensional $\bd{\Sigma}(\bd{U})$. Specifically, we replace the kernel weights used in \cite{chen2016dynamic} with weights derived from ``honest forests'', which are determined by the proportion of trees in which an observation shares the same leaf as the target covariate vector \citep{wager2018estimation, athey2019generalized, friedberg2020local}.   
The true nonparametric covariance matrix is decomposed into two components, as outlined in \citep{chen2016dynamic, chen2019new}. For each component, we construct random forest trees using a specially designed splitting rule aimed at maximizing the difference between the estimators of the child nodes. Given a covariate vector $\bd{u}$, the trained random forest identifies the nearest neighbor observations from the out-of-bag data---data that is not used during the tree-growing process. These out-of-bag nearest neighbors are then used to estimate the conditional covariance matrix at $\bd{U} = \bd{u}$.   
Compared to traditional kernel and nearest-neighbor methods, which rely on a weighted average of nearby data points based on the distance of covariate vectors to $\bd{u}$, honest forests take a more data-driven approach. They determine the importance of nearby observations by accounting for the relationship between covariates and the response. This feature is particularly advantageous in scenarios with multiple conditioning covariates, where traditional methods struggle with the ``curse of dimensionality'' \citep{wager2018estimation}.


We make several key contributions in this paper. To the best of our knowledge, this is the first study to address the estimation of high-dimensional dynamic covariance matrices with multiple conditioning covariates within a nonparametric modeling framework, where the dimensionality of the response may exceed the sample size.
A significant theoretical contribution of our work is demonstrating that the convergence rate of the estimated covariance matrices, derived using the proposed approach, holds uniformly over a compact set of dynamic covariates of interest. This result is valid under the condition that the dimensionality of the response grows sub-exponentially with the sample size, as the sample size tends to infinity. 
Our findings extend the scope of prior research, which focused on cases with a single conditioning covariate \citep{chen2016dynamic}, by addressing the more challenging and practically relevant setting of multiple conditioning covariates in high-dimensional scenarios.

\subsection{Related work}

Numerous studies have explored the estimation of sparse covariance matrices in high dimensions under the assumption that these matrices are static \citep{Bickel_2008, xue2012positive, cai2016estimating, balakrishnan2017computationally}. A fundamental assumption in these works is that the replicated high-dimensional random vectors follow the same probability distribution. However, this assumption of homogeneity often does not hold in practice \citep{chen2016dynamic, JMLR:v23:21-0795, jiang2024dynamic}, necessitating methods that account for heterogeneity.

The estimation of high-dimensional dynamic covariance matrices, where the entries vary in response to certain conditioning covariates, is increasingly in demand. Existing methods for dynamically modeling covariance matrices typically adopt parametric or semi-parametric approaches, relying on structural assumptions about the true covariance models \citep{engle2017large, guo2017dynamic, chen2019new, ZHANG2024110017}. However, such assumptions may not hold in real-world applications, making these methods sensitive to model misspecification. 
Nonparametric methods, on the other hand, are well-known for their robustness to model misspecification. \cite{Chen_2013, chen2016dynamic, wang2021nonparametric} introduced high-dimensional nonparametric covariance models and proposed kernel-smoothing techniques for estimation. While these methods are theoretically appealing, their performance deteriorates in finite samples when the number of conditioning covariates is moderately large (or even as small as three) due to the ``curse of dimensionality" inherent in kernel-smoothing techniques. Notably, these studies only addressed cases with a single conditioning covariate.

Recent advancements in machine learning have provided promising alternatives. \citet{wager2018estimation} introduced the ``honest forest". The honest forest can take both covariates and response variables into account during the branching process, which is especially crucial in scenarios with multiple conditioning covariates. \citet{wager2018estimation} demonstrated that the bias of honest forests converges to zero, and \citet{athey2019generalized} established their consistency properties.  Building on these ideas, \citet{alakus2023covariance} applied honest forests to the estimation of covariance matrices. However, their work is limited to cases where the dimensionality of the response is fixed, and they do not provide theoretical guarantees for high-dimensional settings. 
To the best of our knowledge, there has been no theoretical extension of the honest forest to settings where the dimensionality of the response exceeds the sample size. Our work addresses this critical gap by establishing the consistency rate of honest forests in such settings, showing that the approach remains valid when the dimensionality of the response grows sub-exponentially with the sample size as the sample size tends to infinity.

\subsection{Notation} We introduce some  notation. Define $[n]=\{1,2,\dots, n\}$ for any positive integer $n$.
Let $\mu_{\max}(\cdot)$, $\mu_{\min}(\cdot)$ and $\mu_j(\cdot)$ be the largest, smallest and $j$-th largest eigenvalues 
of an input matrix, respectively.
For a matrix $\mb A=[a_{jr}]_{p\times q}$, we denote the spectral norm $\|\mb A\|_2=\sqrt{\mu_{\max}(\mb A\mb A^\top)}$, the max norm $\|\mb A\|_{\max}=\max_{j,r}|a_{jr}|$, and the Frobenius norm $\|\mb A\|_F=\sqrt{\operatorname{trace}(\mb A\mb A^\top)}=\sqrt{\sum_{j=1}^p\mu_j(\mb A\mb A^\top)}$. Let $\mb I_{p\times p}$ be the $p\times p$ identity matrix. For a vector $\bd v=(v_1,\ldots,v_{d_v})^{\top}\in \mathbb{R}^{d_v}$, we denote the $L_{\infty}$ norm $\|\bd v\|_{\infty}=\max_{i\in [d_v]}|v_i|$, and the $L_2$ norm $\|\bd v\|_2=\sqrt{\sum_{i=1}^{d_v}v_i^2}$. Let $\bd 1_{p}\in \mathbb{R}^p$ be an column vector of ones. For a set $\mathcal{S}$, let $|\mathcal{S}|$ denote its cardinality. For any scalar $a$, define $\lfloor a \rfloor$ as the floor of $a$ and $\lceil a \rceil$ as the ceil of $a$. For two positive sequences \( a_n \) and \( b_n \), we write \( a_n \asymp b_n \) if both \( a_n/b_n \) and \( b_n/a_n \) are bounded, $a_n \ll b_n$ if $a_n/b_n\rightarrow 0$ as $n\rightarrow \infty$ and $a_n=\mathcal{O}(b_n)$ if $\exists M>0$ such that $a_n\leq Mb_n$ for sufficiently large $n$. For a sequence of random variables $X_n$ and a positive sequence $a_n$, we write $X_n=\mathcal{O}_p(a_n)$ if $\lim_{M\rightarrow \infty}\sup_n P(|X_n|>Ma_n)=0$. Let $\mathbb{I}\{\cdot\}$ be the indicator function. Define $\mbox{vec}(\cdot)$ as an operator which stacks the entries of a matrix into a column vector. Specifically,  for a matrix $\mb A$, the entry $a_{jr}$ corresponds to the $k$-th entry of $\mbox{vec}(\mb A)$ with $k = j + (r-1)p$. 

The remainder of the article is organized as follows. In 
Section \ref{The Model and Methodology}, we detail our proposed estimation method, which combines random forests with thresholding rules for dynamic covariance matrix estimation.
In Section \ref{Theory}, we establish the theoretical properties of our estimators, including the uniform consistency of the dynamic covariance matrix estimators, the consistency of the sparse pattern estimation, and the uniform consistency of the inverse of the estimated dynamic covariance matrices.
In Sections \ref{Simulation Studies} and \ref{Real Data}, we present numerical simulations and apply our method to a case study using stock data from the New York Stock Exchange, respectively. In 
Section \ref{Conclusion}, we summarize our findings.
Additionally, all proofs, detailed algorithms, supplementary simulation results, and additional real data analysis results are provided in the Appendix.

\section{Methodology}
\label{The Model and Methodology}

Conditional on \(\bd U = \bd u\), the conditional covariance matrix can be expressed as 
$
\bd \Sigma(\bd u) = \mathbb{E}(\bd Y \bd Y^\top | \bd U = \bd u) - \mathbb{E}(\bd Y | \bd U = \bd u) \mathbb{E}(\bd Y | \bd U = \bd u)^\top
$ 
\citep{chen2016dynamic}. To estimate \(\bd \Sigma(\bd u)\), we first focus on estimating \(\mathbb{E}(\bd Y \bd Y^\top | \bd U = \bd u)\). 
The matrix \(\mathbb{E}(\bd Y \bd Y^\top | \bd U = \bd u)\) can be restructured by extracting each column and concatenating them into a single column vector. Specifically, we define 
$
\bd Y \bd Y^\top = (\widetilde{\bd Y}^{(1)}, \ldots, \widetilde{\bd Y}^{(p)}),
$
where \(\widetilde{\bd Y}^{(j)} := \bd Y \cdot Y_j\) represents the \(j\)-th column of \(\bd Y \bd Y^\top\). By stacking these columns, we obtain 
$
\widetilde{\bd Y} = (\widetilde{\bd Y}^{(1)\top}, \ldots, \widetilde{\bd Y}^{(p)\top})^\top.
$
The target parameter \(\bd m_1(\bd U) = \mathbb{E}(\widetilde{\bd Y} | \bd U)\) represents the conditional mean of \(\widetilde{\bd Y}\) given \(\bd U\). 
To estimate \(\bd m_1(\bd u)\), we solve the local estimating equation 
$
\mathbb{E}\left(\bd \psi_{\bd m_1(\bd u)}(\widetilde{\bd Y}) | \bd U = \bd u\right) = 0,
$
where:
$
\bd \psi_{\bd m_1(\bd u)}(\widetilde{\bd Y}) = \widetilde{\bd Y} - \bd m_1(\bd u).
$ 
The solution to this equation is obtained by incorporating similarity weights \(\beta_i(\bd u)\), which quantify the relevance of the \(i\)-th training sample in approximating \(\bd m_1(\cdot)\) at \(\bd u\). Using these weights, we estimate the target quantity by applying an empirical version of the estimating equation:
\begin{equation}\label{m_2hat}
\widehat{\bd m}_1(\bd u) = \argmin_{\bd m_1(\bd u)} \left\| \sum_{i=1}^n \beta_i(\bd u) \bd \psi_{\bd m_1(\bd u)}(\widetilde{\bd Y}_i) \right\|_2 = \sum_{i=1}^n \beta_i(\bd u) \widetilde{\bd Y}_i.
\end{equation}

In this paper, we aim to leverage honest forest-based algorithms to adaptively learn problem-specific weights, \(\beta_i(\bd u)\), which can be effectively incorporated into (\ref{m_2hat}) \citep{athey2019generalized}. These weights are designed to capture the similarity between the \(i\)-th training sample and the target value of the covariate vector \(\bd u\), enabling more refined and adaptive estimation. 
Specifically, we begin by growing a collection of \(B\) trees, indexed by \(b \in [B]\). For each tree, we define \(L_b^{\beta}(\bd u)\) as the set of training examples that fall into the same leaf as \(\bd u\). The weights \(\beta_i(\bd u)\) are then constructed to represent the frequency with which the \(i\)-th training example lands in the same leaf as \(\bd u\) across all trees in the forest. Formally, the weights are given by:
\begin{equation} \label{weight}
\beta_i(\bd u) = \frac{1}{B} \sum_{b=1}^B \beta_{bi}(\bd u), \quad \text{where} \quad \beta_{bi}(\bd u) = \frac{\mathbb{I}\{\bd U_i \in L_b^{\beta}(\bd u)\}}{|L_b^{\beta}(\bd u)|}.
\end{equation} 
These weights \(\beta_i(\bd u)\), defined for \(i \in [n]\), sum to 1 and reflect the degree of similarity between the \(i\)-th observation and the covariate vector \(\bd u\). The details of estimating the weights are provided in Algorithm \ref{alg: Generalized random forest with honesty and subsampling}. A higher weight indicates a greater similarity between the two.  
To assess similarity, we use a random forest, where each decision tree partitions the feature space into distinct hyperrectangles, clustering observations within the same leaf as similar. The overall similarity between two samples increases with the frequency they share leaf nodes across the forest. This approach, conceptually related to \(k\)-PNN \citep{wager2018estimation}, enables our method to adaptively capture local data relationships, enhancing both performance and robustness.

\begin{algorithm}[t]
\setstretch{1}
\caption{Estimation of weight vector $\bd \beta (\bd u)=(\beta_1(\bd u),\ldots,\beta_n(\bd u))^{\top}$}\label{alg: Generalized random forest with honesty and subsampling}
\begin{algorithmic}[1]
    \State \textbf{Given:} Training dataset $\{(\bd Y_i,\bd U_i)\}_{i=1}^n$, target point $\bd u$, the number of trees $B$ and the subsample size $s$ used in \Call{Subsample}{}
    \Function{WeightVectorEstimation$_\beta$}{$\{(\bd Y_i,\bd U_i)\}_{i=1}^n,\bd u,s$}
        \State \textbf{Initialize} weight vector $\bd \beta(\bd u)$ as the zero vector $\mathbf{0}_{n}$ and data matrix $\widetilde{\mb Y}$ as the zero matrix $\mb O_{p^2 \times n}$
        \For{$i\gets 1$ to $n$}
        \State $\widetilde{\mb Y}[,i]\gets$ \Call{Vec}{$\bd Y_i\bd Y_i^{\top}$} \Comment{$\widetilde{\mb Y}[,i]$ is the $i$-th column of $\widetilde{\mb Y}$.}
        \EndFor
        \For{$b\gets 1$ to total number of trees $B$}
		\State set of samples $\mathcal{I}\gets$ \Call{Subsample}{$\{(\widetilde{\mb Y}[,i],\bd U_i)\}_{i=1}^n,s$}
            \State set of samples $\mathcal{J}_1$, $\mathcal{J}_2\gets$ \Call{Splitsample}{$\mathcal{I}$}
            \State index $I_{\mathcal{J}_2}\gets \{i:(\bd Y_i,\bd U_i)\in \mathcal{J}_2\}$
            \State tree $\mathcal{T}\gets$ \Call{GradientTree}{$\mathcal{J}_1$,$\{\bd U_i\}_{i\in I_{\mathcal{J}_2}}$} \Comment{See Algorithm S.2 in Supplementary Materials S.3.}
            \State $\mathcal{N}\gets$ \Call{Neighbors}{$\bd u$, $\mathcal{T}$, $\mathcal{J}_2$} \Comment{Returns indexes of elements of $\mathcal{J}_2$ that fall into \\ ~~~~~~~~~~~~~~~~~~~~~~~~~~~~~~~~~~~~~~~~~~~~~~~~~~ the same leaf as $\bd u$ in the tree $\mathcal{T}$.}
            \For{$j\in \mathcal{N}$}
                \State $\bd \beta(\bd u)[j]$ += $1/(B|\mathcal{N}|)$ \Comment{$\bd \beta(\bd u)[j]$ is the $j$-th element of $\bd \beta(\bd u)$.}
            \EndFor
        \EndFor
        \State \textbf{Return} weight vector $\bd \beta(\bd u)$.
    \EndFunction
    \State The function \Call{Vec}{} stacks the entries of a matrix into a column vector; \Call{Subsample}{} draws a subsample of size $s$ from $\{(\widetilde{\mb Y}[,i],\bd U_i)\}_{i=1}^n$ without replacement; and \Call{Splitsample}{} randomly divides a set into two evenly sized and nonoverlapping halves.
\end{algorithmic}
\end{algorithm}


We next discuss how to grow trees. Given a parent node \(\mathcal{P}\), we  divide it into two child nodes, \(\mathcal{C}_1\) and \(\mathcal{C}_2\), by applying an axis-aligned cut that maximizes the heterogeneity of the in-sample \(\bd{m}_1\)-estimates as effectively as possible \citep{athey2019generalized}. Specifically, 
define
\begin{equation}\label{mhat2P}
    \widehat{\bd m}_{1,\mathcal{C}_1}=\frac{1}{n_{\mathcal{C}_1}}\sum_{\{ i:\bd U_i\in \mathcal{C}_1 \}}\widetilde{\bd Y}_i,
\end{equation} where $n_{\mathcal{C}_1}=|\{ i:\bd U_i\in\mathcal{C}_1 \}|$.  Define $\widehat{\bd m}_{1,\mathcal{C}_2}$ and $n_{\mathcal{C}_2}$ similarly. Then we define 
\begin{equation}\label{Delta2}
    \Delta_1(\mathcal{C}_1,\mathcal{C}_2)=\|\widehat{\bd m}_{1,\mathcal{C}_1}-\widehat{\bd m}_{1,\mathcal{C}_2}\|_2^2n_{\mathcal{C}_1}n_{\mathcal{C}_2}/n_{\mathcal{P}}^2.
\end{equation}  
We split \(\mathcal{P}\) into two axis-aligned child nodes, \(\mathcal{C}_1\) and \(\mathcal{C}_2\), in order to maximize the criterion \(\Delta_1(\mathcal{C}_1, \mathcal{C}_2)\). After dividing \(\mathcal{P}\) into these two children, we then treat each child as a new parent node and recursively repeat the process. The detailed algorithm can be found in Algorithm S.2 in Supplementary Materials S.3.
 This data-driven partitioning approach takes into account the relationship between covariates and the response vector. Partitioning is based on covariates that significantly impact the response vector, as they create notable differences in estimators across the child nodes. This method is especially well-suited for cases where only certain components of the covariate vector influence the response vector.
 
 Our approach builds  on two key concepts: training trees on subsamples of the training data \citep{scornet2015consistency, wager2018estimation} and  utilizing a subsample-splitting method known as honesty \citep{biau2012analysis, wager2018estimation}. A tree (the $b$-th tree) is considered honest if each response $\widetilde{\bd Y}_i$  is used either to decide where to place the splits  or to estimate  $\bd m_1(\bd u)$ within the leaf, but not for both tasks. Our algorithm implements honesty by splitting the training subsample into two parts, $\mathcal{J}_1$ and $\mathcal{J}_2$. The $\mathcal{J}_1$ sample and $\bd U$-observations from the $\mathcal{J}_2$ sample are used to determine splits, while the $\mathcal{J}_2$ sample is reserved to estimate $\bd m_1(\bd u)$ via $\sum_{\{i:(\widetilde{\bd Y}_i,\bd U_i) \in  \mathcal{J}_2\}}\beta_{bi}(\bd u)\widetilde{\bd Y}_i$. 
Note that, we ensure that the proportion of  the $\mathcal{J}_2$ sample in each ``branch" does not become too small when making a split, and further, each leaf of the tree contains at least $k$ $\mathcal{J}_2$ observations, as described in Algorithm S.2 in Supplementary Materials S.3. 

We can similarly express the forest-based estimator of $\bd m(\bd u) = \mathbb{E}(\bd Y|\bd U = \bd u)$ as follows:
\begin{equation}\label{mhat}
    \widehat{\bd m}(\bd u)=\sum_{i=1}^n\alpha_i(\bd u)\bd Y_i.
\end{equation}
The function used to obtain the weights $\{\alpha_i(\bd u)\}_{i=1}^n$, referred to as {\it WeightVectorEstimation$_\alpha$}, is deferred to Appendix S.3 and presented in Algorithm S.1.
 Specifically, 
we  grow  \( B \) trees. For each tree, we define \( L_b^{\alpha}(\bd u) \) as the set of training examples that fall into the same ``leaf" as \( \bd u \). The weights \( \alpha_i(\bd u) \) are define as 
$
   \alpha_i(\bd u) = {B}^{-1} \sum_{b=1}^B \alpha_{bi}(\bd u)$ with $ \alpha_{bi}(\bd u) = \frac{\mathbb{I}\{\bd U_i \in L_b^{\alpha}(\bd u)\}}{|L_b^{\alpha}(\bd u)|}.
$
Combining  (\ref{m_2hat}) and (\ref{mhat}), we can obtain an estimator of  the conditional covariance matrix:
\begin{equation}
    \widehat{\mb \Sigma}(\bd u)=\sum_{i=1}^n\beta_i(\bd u)\bd Y_i\bd Y_i^\top - \left\{\sum_{i=1}^n\alpha_i(\bd u)\bd Y_i\right\}\left\{\sum_{i=1}^n\alpha_i(\bd u)\bd Y_i^\top\right\}. \label{covariancehat}
\end{equation}
It's noted that the $\beta_i$ in (\ref{covariancehat}) is the same as the $\beta_i$ in (\ref{m_2hat}).
In high-dimensional settings where the number of variables \( p \) greatly exceeds the sample size \( n \), \( \widehat{\mb \Sigma}(\bd u) \) becomes singular, making it unsuitable for estimating the inverse of a covariance matrix. With the growing availability of large datasets, there is a strong demand for new methods that can effectively estimate dynamic covariance matrices while possessing desirable theoretical properties.




We propose integrating the generalized threshold operator \citep{rothman2009generalized}, including methods such as hard thresholding \citep{Bickel_2008}, Lasso \citep{tibshirani1996regression}, adaptive Lasso \citep{zou2006adaptive}, and SCAD \citep{fan2001variable}, into \( \widehat{\mb \Sigma}(\bd u) \).
The generalized shrinkage operator for all $z\in\mathbb{R}$ is denoted as the function $s_{\lambda}: \mathbb{R}\rightarrow\mathbb{R}$, which satisfies three conditions: (i) $|s_{\lambda}(z)|\leq|z|$; (ii) $s_{\lambda}(z)=0$ for $|z|\leq \lambda$; and  (iii) $|s_{\lambda}(z)-z|\leq\lambda$. 
We  construct our estimator as
\begin{equation*}
    s_{\lambda(\bd u)}(\widehat{\mb \Sigma}(\bd u))=\left[s_{\lambda(\bd u)}(\widehat{\sigma}_{jr}(\bd u))\mathbb{I}\{j\neq r\}+\widehat{\sigma}_{jr}(\bd u)\mathbb{I}\{j=r\}\right]_{p\times p},
\end{equation*}
where $\lambda(\bd u)$ is a dynamic thresholding parameter depending on $\bd u$. It is noted that no penalty is imposed on the diagonal elements of $\widehat{\mb \Sigma}(\bd u)$. We refer to our covariance estimates as Forest-based Dynamic Covariance Models (FDCMs) to highlight both the forest-based estimator and the dependence of the conditional mean and conditional covariance matrix on the dynamic index vector \(\bd U\). Notably, we allow the thresholding parameter \(\lambda\) to vary with the dynamic vector \(\bd U\). Consequently, the proposed FDCMs can fully adapt to the varying sparsity levels associated with different \(\bd U\).


\section{Theory}
\label{Theory}

Throughout this article, we implicitly assume $n \ll p$. It is essential to clarify that, in the case of $n \ll p$, due to the insufficient sample size, it is almost impossible to provide well-behaved estimators if all $p^2$ elements in the covariance matrix are unrestricted non-zero elements. Therefore, we need to impose some constraints on  the dynamic covariance matrices. 
Specifically,  we define the set of covariance matrices  as:
\begin{small}
\begin{equation*}
    \mathcal{U}(0,c_0(p),M_1;\Omega)
    =\left\{\{\bd \Sigma(\bd u),\bd u\in\Omega\}\middle|\sup_{\bd u\in\Omega}\sigma_{jj}(\bd u)<M_1<\infty, \sup_{\bd u\in\Omega}\left(\sum_{r=1}^p\mathbb{I}\{\sigma_{jr}(\bd u)\neq0\}\right)\leq c_0(p),\forall j\right\},
\end{equation*}
\end{small}%
where $\sigma_{jr}(\bd u)$ represents the element in the $j$-th row and $r$-th column of $\bd \Sigma(\bd u)$, and $\Omega$ is a compact subset of $\mathbb{R}^d$. We can extend this definition to
\begin{small}
\begin{equation*}
    \mathcal{U}(q,c_0(p),M_1;\Omega)
    =\left\{\{\bd \Sigma(\bd u),\bd u\in\Omega\}\middle|\sup_{\bd u\in\Omega}\sigma_{jj}(\bd u)<M_1<\infty,\sup_{\bd u\in\Omega}\left(\sum_{r=1}^p|\sigma_{jr}(\bd u)|^q\right)\leq c_0(p),\forall j\right\},
\end{equation*}
\end{small}%
where $0\leq q<1$. This definition is identical to the conditions specified for the dynamic covariance matrices in \citet{chen2016dynamic}. We also require the following regularity conditions, which are mild, and their justification will be provided in Supplementary Materials S.1.  

\subsection*{\textbf{Regularity Conditions:}}

\begin{enumerate}
    \item[(a)]  
    The random vectors \(\bd U_1, \dots, \bd U_n\) are assumed to be independent and identically distributed, each following a uniform distribution on \([0,1]^d\),  denoted as \(\operatorname{Unif}([0,1]^d)\). 
    \item[(b)] The conditional first and second moments of $\bd Y$ are both Lipschitz continuous in $\bd u$. Specifically,  $\forall \bd u_1, \bd u_2 \in [0,1]^d$, it holds that
    \begin{eqnarray*}
        &&\|\mbox{vec}\left(\mathbb{E}(\bd Y\bd Y^{\top}|\bd U=\bd u_1)\right)-\mbox{vec}\left(\mathbb{E}(\bd Y\bd Y^{\top}|\bd U=\bd u_2)\right)\|_{\infty}\leq C_L\|\bd u_1-\bd u_2\|_{\infty}, 
\\
        &&\|\mathbb{E}(\bd Y|\bd U=\bd u_1)-\mathbb{E}(\bd Y|\bd U=\bd u_2)\|_{\infty}\leq C_L\|\bd u_1-\bd u_2\|_{\infty},
    \end{eqnarray*}
    where $C_L$ is the Lipschitz constant.  
    \item[(c)] Each element in  $\bd Y_i\bd Y_i^{\top}$ and $\bd Y_i$, given $\bd U_i=\bd u$,  $\forall \bd u\in[0,1]^d$, satisfies the Bernstein's condition, i.e. $\forall i\in[n]$ and $\forall j,r\in[p]$, the following inequalities hold:
    \begin{equation*}
        \sup_{\bd u\in[0,1]^d}\left|\mathbb{E}\left(\{Y_{ij}Y_{ir}-\mathbb{E}(Y_{ij}Y_{ir}|\bd U_i=\bd u)\}^l\middle|\bd U_i=\bd u\right)\right|\leq \frac{1}{2}l!\sigma_B^2m^{l-2},  ~~\mbox{ and}
    \end{equation*}
    \begin{equation*}
        \sup_{\bd u\in[0,1]^d}\left|\mathbb{E}\left(\{Y_{ij}-\mathbb{E}(Y_{ij}|\bd U_i=\bd u)\}^l\middle|\bd U_i=\bd u\right)\right|\leq \frac{1}{2}l!\sigma_B^2m^{l-2}, \; \forall l\geq2,
    \end{equation*}
    where $m<\infty$ is the parameter of the Bernstein's condition and $\sigma_B^2$ is defined as
    \begin{equation*}
    \begin{split}
    \sigma_B^2:=\max\left\{\sup_{\bd u \in [0,1]^d}\max_{j,r}\mbox{Var}(Y_{ij}Y_{ir}|\bd U_i=\bd u), \sup_{\bd u \in [0,1]^d}\max_{j}\mbox{Var}(Y_{ij}|\bd U_i=\bd u)\right\}<\infty.
    \end{split}
    \end{equation*}
    \item[(d)] For the $b$-th tree, $b\in[B]$, it  holds that
 $   
        \mathbb{E}(\sup_{\bd u\in[0,1]^d} \|\sum_{i=1}^n\beta_{bi}(\bd u)\widetilde{\bd Y}_i \|_{\infty})<\infty$  and 
$        \mathbb{E}(\sup_{\bd u\in[0,1]^d} \|\sum_{i=1}^n\alpha_{bi}(\bd u)\bd Y_i \|_{\infty})<\infty, $
    where $\widetilde{\bd Y}_i=\mbox{vec}(\bd Y_i\bd Y_i^{\top})$.
    \item[(e)] 
    All trees are symmetric, meaning their output remains invariant under any permutation of the indices of the training examples. The trees are also \(\omega\)-regular, for some \(0 < \omega \leq 0.2\), ensuring that every split allocates at least a fraction \(\omega\) of the observations from the parent node to each child node. Additionally, each terminal node contains between \(k\) and \(2k-1\) observations, for some \(k \in \mathbb{N}\).
The trees are constructed as random-split trees, where at every split, the probability of selecting the \(j\)-th feature as the splitting variable is bounded below by \(\pi/d\), with \(0 < \pi \leq 1\) and \(j \in [d]\). 
The forest is honest and built using subsampling, with the subsample size \(s\) satisfying \(s/n \to 0\) and \(s \to \infty\). This ensures the robustness and asymptotic properties of the resulting estimates.

\end{enumerate}

We next establish the consistency of our proposed estimator,  $s_{\lambda(\bd u)}(\widehat{\mb \Sigma}(\bd u))$, in terms of the max norm, the spectral norm and the Frobenius norm. 

\begin{theorem} [Uniform consistency of the estimated dynamic matrix]
\label{Uniform consistency in estimation}
    Under Conditions (a)--(e), suppose that $s_{\lambda}$ is a shrinkage operator. Uniformly on $\mathcal{U}(q,c_0(p),M_1;[0,1]^d)$, if $\lambda_n(\bd u)=M(\bd u)\left(\sqrt{s\left\{\log(n)+\log(p)\right\}/n}+s^{-\frac{1}{2}\frac{\log(1-\omega)}{\log(\omega)}\frac{\pi}{d}}\right)$ and $s\log(p)/n\rightarrow0$, we have
    \begin{equation*}
    \begin{aligned}
        &\sup_{\bd u\in[0,1]^d}\|s_{\lambda_n(\bd u)}(\widehat{\mb \Sigma} (\bd u))-\mb \Sigma(\bd u)\|_{\max}=\mathcal{O}_p\left(\sqrt{{s\left\{\log(n)+\log(p)\right\}}/{n}}+s^{-\frac{1}{2}\frac{\log(1-\omega)}{\log(\omega)}\frac{\pi}{d}}\right), \\
        &\sup_{\bd u\in[0,1]^d}\|s_{\lambda_n(\bd u)}(\widehat{\mb \Sigma} (\bd u))-\mb \Sigma(\bd u)\|_2=\mathcal{O}_p\left(c_0(p)\left(\sqrt{{s\left\{\log(n)+\log(p)\right\}}/{n}}+s^{-\frac{1}{2}\frac{\log(1-\omega)}{\log(\omega)}\frac{\pi}{d}}\right)^{1-q}\right), \\
        &{p}^{-1}\sup_{\bd u\in[0,1]^d}\|s_{\lambda_n(\bd u)}(\widehat{\mb \Sigma}(\bd u))-\mb \Sigma(\bd u)\|_F^2=\mathcal{O}_p\left(c_0(p)\left(\sqrt{{s\left\{\log(n)+\log(p)\right\}}/{n}}+s^{-\frac{1}{2}\frac{\log(1-\omega)}{\log(\omega)}\frac{\pi}{d}}\right)^{2-q}\right),
    \end{aligned}
    \end{equation*}
    where $M(\bd u)$ depending on $\bd u\in[0,1]^d$ is large enough and $\sup_{\bd u\in[0,1]^d}M(\bd u)<\infty$.
\end{theorem}

Theorem \ref{Uniform consistency in estimation} demonstrates the consistency of our estimator under the max norm, the spectral norm and the Frobenius norm. The detailed proofs of Theorems \ref{Uniform consistency in estimation}  can be found in Supplementary Materials S.2. 
We observe that the bias is bounded uniformly by $\mathcal{O}\left(s^{-\frac{1}{2}\frac{\log(1-\omega)}{\log(\omega)}\frac{\pi}{d}}\right)$ and the variance is of the order $\mathcal{O}_p\left(\sqrt{{s}{n}^{-1}\{\log(n)+\log(p)\}}\right)$ uniformly. 
Compared to the results in \cite{wager2018estimation}, which are limited to a scalar response and do not provide uniform convergence results, our estimator exhibits the same convergence rate for the bias term, but the convergence rate of the variance term includes additional logarithmic factors, specifically $\log(n)$ and $\log(p)$. The $\log(n)$ term corresponds to the complexity involved in achieving a uniform bound over a compact set, whereas the $\log(p)$ term arises due to the effects of working in high-dimensional spaces. Most importantly, our result holds consistently across a range of conditioning  covariates used for modeling dynamics. This is the first uniform result that combines the strengths of random forests and covariance matrix estimation in the challenging high-dimensional regime. Moreover, 
it is important to highlight that the original work on generalized random forests by \citet{athey2019generalized} focuses solely on point-wise convergence results. In contrast, our study is the first to establish uniform consistency for generalized random forests in regression settings.   


The uniform convergence rate in Theorem \ref{Uniform consistency in estimation} reflects the familiar bias-variance trade-off in the nonparametric estimation literature (Wand and Jones 1995; Fan and Gijbels 1996; Chen \& Leng, 2016; Lu, Kolar \& Liu, 2018). Specifically, for each individual tree, a larger subsample size $s$ allows the tree to grow deeper and perform more splits. This reduces sample heterogeneity within each leaf node, leading to a smaller bias for individual trees. Since the bias of a random forest equals that of its individual trees, increasing $s$ reduces the overall bias.
However, the variance of a random forest depends on both the variance of individual trees and the covariance between trees. As the number of trees approaches infinity, the covariance becomes the dominant factor \citep{genuer2012variance}. When $s$ increases, the differences among trees diminish, leading to higher covariances. Consequently, the variance of the random forest increases with $s$.
Consider the case where \( n \ll p \). By setting \( s \asymp \left(n/\log (p)\right)^{\frac{d\log (\omega)}{d\log (\omega)+\pi\log(1-\omega)}} \) to balance bias and variance, we achieve the optimal convergence rate \( \left(\log (p)/n\right)^{\frac{\pi\log(1-\omega)}{2(d\log (\omega)+\pi\log(1-\omega))}} \). Therefore, the dimension $p$ can be of the order $o(\exp(n))$ to ensure a meaningful convergent estimator.
Since our estimating method is nonparametric, its convergence rate is slower than the conventional parametric rate of \( \left(\log (p)/n\right)^{\frac{1}{2}} \) as discussed by \citet{Bickel_2008}.

When dealing with sparse covariance matrices, it is crucial to evaluate whether the estimator accurately captures the sparsity pattern of the true covariance matrix. In this context, we provide theoretical guarantees.

\begin{theorem}[Uniform consistency in estimating the sparsity pattern]\label{Uniform consistency in estimating the sparsity pattern}
    Under Conditions (a)--(e), suppose that $s_{\lambda}$ is a shrinkage operator, and that $0<\sigma_{jj}(\bd u)<M_1<\infty$, $\forall j\in [p]$ and $\bd u\in [0,1]^d$. If $\lambda_n(\bd u)=M(\bd u)\left(\sqrt{s\left\{\log(n)+\log(p)\right\}/n}+s^{-\frac{1}{2}\frac{\log(1-\omega)}{\log(\omega)}\frac{\pi}{d}}\right)$ with $M(\bd u)$ depending on $\bd u\in[0,1]^d$ large enough and satisfying $\sup_{\bd u\in[0,1]^d}M(\bd u)<\infty$ and  $s\log(p)/n\rightarrow0$, we have
 $   
        s_{\lambda_n(\bd u)}(\widehat{\sigma}_{jr}(\bd u))=0\textit{ for }\sigma_{jr}(\bd u)=0,\;\forall(j,r) \textit{ satisfying } j\neq r, 
$  
    with probability tending to $1$ uniformly in $\bd u\in[0,1]^d$. We assume further that, for each $\bd u\in[0,1]^d$, all nonzero elements of $\mb \Sigma(\bd u)$ satisfy $|\sigma_{jr}(\bd u)|>\tau_n(\bd u)$, where
    \begin{equation}\label{sparsity_condition}
       \left[ {\sqrt{{s\left\{\log(n)+\log(p)\right\}}/{n}}+s^{-\frac{1}{2}\frac{\log(1-\omega)}{\log(\omega)}\frac{\pi}{d}}}\right]/{\inf_{\bd u\in[0,1]^d}(\tau_n(\bd u)-\lambda_n(\bd u))}\rightarrow0.
    \end{equation} Then, we have 
   $
        \operatorname{sign}\{(s_{\lambda(\bd u)}(\widehat{\sigma}_{jr}(\bd u))\mathbb{I}\{j\neq r\}+\widehat{\sigma}_{jr}(\bd u)\mathbb{I}\{j=r\})\cdot\sigma_{jr}(\bd u)\}=1$  for $\sigma_{jr}(\bd u)\neq0,\;\forall(j,r),
$     
    with probability tending to $1$ uniformly on $\bd u\in[0,1]^d$.
\end{theorem}

    The conditions $|\sigma_{jr}(\bd u)|>\tau_n(\bd u)$ and (\ref{sparsity_condition}) ensure that the nonzero elements of $\bd \Sigma(\bd u)$ can be distinguished from the noise and have also been used in \cite{chen2016dynamic}. The condition (\ref{sparsity_condition}) holds if  $\inf_{\bd u\in[0,1]^d}\tau_n(\bd u)$ satisfies:  $\inf_{\bd u\in[0,1]^d}\tau_n(\bd u)=M_n( \sqrt{s\left\{\log(n)+\log(p)\right\}/n} \\+s^{-\frac{1}{2}\frac{\log(1-\omega)}{\log(\omega)}\frac{\pi}{d}} ),$ where $M_n\rightarrow \infty$ at an arbitrarily slow rate. 
Proof details of Theorem \ref{Uniform consistency in estimating the sparsity pattern} are provided in  Supplementary Materials S.2. This theorem demonstrates that our estimator can effectively distinguish between zero and significantly non-zero entries in the dynamic covariance matrices over $\bd u\in[0,1]^d$.

Define $\mathcal{U}(q,c_0(p),M_1,\epsilon;\Omega)$ as 
\begin{equation*}
\big\{\{\mb \Sigma(\bd u),\bd u\in\Omega\} | \{\mb \Sigma(\bd u),\bd u\in\Omega\}\in\mathcal{U}(q,c_0(p),M_1;\Omega),
    \inf_{\bd u\in \Omega}\left(\mu_{\min}(\mb \Sigma(\bd u))\right)\geq\epsilon>0\big\}, 
\end{equation*}
which is a set consisting of only positive definite dynamic covariances in $\mathcal{U}(q,c_0(p),M_1;\Omega)$. We now provide the uniform convergence rate for the inverse of the covariance matrix estimator. The detailed proof of Theorem \ref{Uniform consistency of the inverse of the estimated dynamic matrix}  can be found in Supplementary Materials S.2.

\begin{theorem}\label{Uniform consistency of the inverse of the estimated dynamic matrix}
    \textbf{\textup{(Uniform consistency of the inverse of the estimated dynamic matrix)}}
    Under Conditions (a)--(e), suppose that $s_{\lambda}$ is a shrinkage operator. If $\lambda_n(\bd u)=M(\bd u)\left(\sqrt{s\left\{\log(n)+\log(p)\right\}/n}+s^{-0.5\frac{\log(1-\omega)}{\log(\omega)}\frac{\pi}{d}}\right)$ and $s\log(p)/n\rightarrow0$ and \\ $c_0(p)\left(\sqrt{s\left(\log(n)+\log(p)\right)/n}+s^{-\frac{1}{2}\frac{\log(1-\omega)}{\log(\omega)}\frac{\pi}{d}}\right)^{1-q}\rightarrow0$, we have that uniformly on \\ $\mathcal{U}(q,c_0(p),M_1,\epsilon;[0,1]^d)$,
    \begin{equation*}
    \begin{aligned}
        &\sup_{\bd u\in[0,1]^d}\left\|\left(s_{\lambda_n(\bd u)}(\widehat{\mb \Sigma}(\bd u))\right)^{-1}-\mb \Sigma^{-1}(\bd u)\right\|_2=\mathcal{O}_p\left(c_0(p)\left(\sqrt{{s\left\{\log(n)+\log(p)\right\}}/{n}}+s^{-0.5\frac{\log(1-\omega)}{\log(\omega)}\frac{\pi}{d}}\right)^{1-q}\right), \\
        &\frac{1}{p}\sup_{\bd u\in[0,1]^d}\left\|\left(s_{\lambda_n(\bd u)}(\widehat{\mb \Sigma}(\bd u))\right)^{-1}-\mb \Sigma^{-1}(\bd u)\right\|_F^2=\mathcal{O}_p\left(c_0(p)\left(\sqrt{{s\left\{\log(n)+\log(p)\right\}}/{n}}+s^{-0.5\frac{\log(1-\omega)}{\log(\omega)}\frac{\pi}{d}}\right)^{2-q}\right),
    \end{aligned}
    \end{equation*}
    where $M(\bd u)$ depending on $\bd u\in[0,1]^d$ is large enough and $\sup_{\bd u\in[0,1]^d}M(\bd u)<\infty$.
\end{theorem}


 To guarantee the positive definiteness of the estimated covariance matrix, we follow the method proposed in \citet{chen2016dynamic}. Specifically, if $s_{\lambda(\bd u)}(\widehat{\mb \Sigma}(\bd u))$ is already positive definite, no such correction is needed. Otherwise, we set $-\widehat{\delta}(\bd u)$ to be the smallest eigenvalue of $s_{\lambda(\bd u)}(\widehat{\mb \Sigma}(\bd u))$ with $\widehat{\delta}(\bd u)\geq0$. We further let $c_n$ be a very small positive number. The corrected estimator is defined as
   $ \widehat{\mb \Sigma}_{C}(\bd u)=s_{\lambda(\bd u)}(\widehat{\mb \Sigma}(\bd u))+\left(\widehat{\delta}(\bd u)+c_n\right)\mb I_{p\times p},$
 The smallest eigenvalue of $\widehat{\mb \Sigma}_{C}(\bd u)$ is $c_n>0$. Therefore, $\widehat{\mb \Sigma}_{C}(\bd u)$ is positive definite.  Taking together, to guarantee positive definiteness, we define a modified estimator of $\mb \Sigma(\bd u)$ as
\begin{equation*}
    \widehat{\mb \Sigma}_M(\bd u)=\widehat{\mb \Sigma}_{C}(\bd u)\mathbb{I}\{\mu_{\min}(s_{\lambda(\bd u)}(\widehat{\mb \Sigma}(\bd u)))\leq0\}+s_{\lambda(\bd u)}(\widehat{\mb \Sigma}(\bd u))\mathbb{I}\{\mu_{\min}(s_{\lambda(\bd u)}(\widehat{\mb \Sigma}(\bd u)))>0\}.
\end{equation*}
The complete estimation process is presented in Algorithm S.7.

Let $c_n=\mathcal{O}\left(c_0(p)\left(\sqrt{s\left\{\log(n)+\log(p)\right\}/n}+s^{-\frac{1}{2}\frac{\log(1-\omega)}{\log(\omega)}\frac{\pi}{d}}\right)^{1-q}\right)$. It can be easily inferred following \citet{chen2016dynamic} that the uniform convergence rate of spectral norm in Theorem \ref{Uniform consistency in estimation} and all the results in Theorem \ref{Uniform consistency in estimating the sparsity pattern} also hold for the modified estimator $\widehat{\mb \Sigma}_M(\bd u)$. The uniform convergence rate of max norm and Frobenius norm in Theorem \ref{Uniform consistency in estimation} does not hold for $\widehat{\mb \Sigma}_M(\bd u)$. This is because  the magnitude of $\widehat{\delta}(\bd u)$ is dominated by $\sup_{\bd u\in[0,1]^d}\|s_{\lambda_n(\bd u)}(\widehat{\mb \Sigma} (\bd u))-\mb \Sigma(\bd u)\|_2$ rather than by $\sup_{\bd u\in[0,1]^d}\|s_{\lambda_n(\bd u)}(\widehat{\mb \Sigma} (\bd u))-\mb \Sigma(\bd u)\|_{\max}$. As a result, the convergence rate becomes slightly slower. 
Specifically, we have
\begin{equation*}
    \begin{aligned}
       \sup_{\bd u\in[0,1]^d}\|\widehat{\mb \Sigma}_M(\bd u)-\mb \Sigma(\bd u)\|_{\max}&\leq \sup_{\bd u\in[0,1]^d}\|s_{\lambda_n(\bd u)}(\widehat{\mb \Sigma} (\bd u))-\mb \Sigma(\bd u)\|_{\max}+|\widehat{\delta}(\bd u)|+c_n \\
       &\leq 2\sup_{\bd u\in[0,1]^d}\|s_{\lambda_n(\bd u)}(\widehat{\mb \Sigma} (\bd u))-\mb \Sigma(\bd u)\|_2+c_n \\
       &=\mathcal{O}_p\left(c_0(p)\left(\sqrt{{s\left\{\log(n)+\log(p)\right\}}/{n}}+s^{-\frac{1}{2}\frac{\log(1-\omega)}{\log(\omega)}\frac{\pi}{d}}\right)^{1-q}\right),
    \end{aligned}
\end{equation*} and
\begin{equation*}
    \begin{aligned}
        {p}^{-1}\sup_{\bd u\in[0,1]^d}\|\widehat{\mb \Sigma}_M(\bd u)-\mb \Sigma(\bd u)\|_F^2&\leq \sup_{\bd u\in[0,1]^d}\|\widehat{\mb \Sigma}_M(\bd u)-\mb \Sigma(\bd u)\|_2^2 \\
        &=\mathcal{O}_p\left(c_0(p)^2\left(\sqrt{{s\left\{\log(n)+\log(p)\right\}}/{n}}+s^{-\frac{1}{2}\frac{\log(1-\omega)}{\log(\omega)}\frac{\pi}{d}}\right)^{2-2q}\right).
    \end{aligned}
\end{equation*} Although $\widehat{\mb \Sigma}_M(\bd u)$ exhibits a slightly slower convergence rate in Frobenius norm than $s_{\lambda_n(\bd u)}(\widehat{\mb \Sigma}(\bd u))$ at the theoretical level, its finite-sample performance closely resembles that of $s_{\lambda_n(\bd u)}(\widehat{\mb \Sigma}(\bd u))$, as demonstrated in Section \ref{Simulation Studies}. 

\section{Simulation Studies}
\label{Simulation Studies}

 In this section, we present numerical simulations to evaluate the performance of the proposed method for estimating large dynamic covariance matrices from finite samples. Specifically, we conduct a series of Monte Carlo experiments and compare our method with alternative approaches, including the static covariance matrix estimators introduced by Rothman, Levina, and Zhu (2009), as well as the kernel-based dynamic covariance matrix (DCM) method proposed by \cite{chen2016dynamic}. These comparisons are performed using various thresholding rules, including the hard, soft, adaptive lasso, and SCAD methods. 
Our simulation studies include both the original and the modified FDCM   approaches. The parameters for the proposed FDCM method encompass those used for training the random forest, as well as the dynamic thresholding parameter \(\lambda(\bd u)\). Details on the selection and tuning of these parameters are provided in Supplementary Materials S.4.


 We consider two dynamic covariance matrix structures: the dynamic AR(1) covariance model and the varying-sparsity covariance model. These structures extend models 2 and 3 from \cite{chen2016dynamic} by incorporating multivariate conditioning covariates. 
In our simulations, we vary the dimension \(p\) of \(\bd Y\) to take values of 100, 150, and 200, while fixing the dimension \(d\) of the conditioning covariates \(\bd U\) at either 10 or 20. The sample size is held constant at \(n = 100\). 
Although \(\bd U\) is a multidimensional vector in our models, only a few components of \(\bd U\) have a significant influence on the covariance matrix. For the kernel-based DCM method, we specifically select one important component and one non-important component of \(\bd U\) for analysis.

For each covariance model, we generate 50 datasets, each consisting of \(n = 100\) observations. The covariate vectors \(\bd U_i, i \in [n]\), are sampled independently and identically from a multivariate uniform distribution over the domain \([ -1, 1]^d\). The response variables \(\bd Y_i\) are drawn from a multivariate normal distribution with a mean of zero and a covariance matrix \(\mb \Sigma(\bd U_i)\).
To evaluate the accuracy of the estimators, we use two metrics: Frobenius loss and spectral loss, which measure the distance between the estimated covariance matrix and the true covariance matrix. Additionally, we generate 30 fixed test points \(\bd u_k, k \in [30]\), sampled from the same multivariate uniform distribution \([ -1, 1]^d\). These test points are held constant across all experiments. 
For each dataset and method, we estimate the covariance matrices at the 30 test points and calculate the median Frobenius loss (MFL) and median spectral loss (MSL) as follows:
\[
\text{Median Frobenius Loss} = \operatorname{median}\left\{\nabla_F(\bd u_k), k \in [30]\right\},
\]
\[
\text{Median Spectral Loss} = \operatorname{median}\left\{\nabla_S(\bd u_k), k \in [30]\right\},
\]
where $
\nabla_F(\bd u) = \|\widehat{\mb \Sigma}(\bd u) - \mb \Sigma(\bd u)\|_F$ and $\nabla_S(\bd u) = \|\widehat{\mb \Sigma}(\bd u) - \mb \Sigma(\bd u)\|_2.$ 
  To assess performance, we compute the average and standard deviation of the MSLs and MFLs over 50 replications for each method.


\subsection*{Dynamic AR(1) Covariance Models:}

\noindent \textbf{Model 1:} Let $\bd u=(u_1,u_2, \ldots ,u_d)^T$ and  $\bd Y \sim N\left(\mathbf{0}, \mb \Sigma(\bd u)\right)$, where $ \mb \Sigma({\bd u})=\left\{\sigma_{j r}({\bd u})\right\}_{1 \leq j, r \leq p}$, $\sigma_{j r}({\bd u})=\exp (u_1) \phi(u_1)^{|j-r|}$ with $\phi(\cdot)$ being the probability density  of $N(0, 1)$.

\noindent \textbf{Model 2:} $\bd Y \sim N\left(\mathbf{0}, \mb \Sigma(\bd u)\right)$, where $\mb \Sigma(\bd u)=\left\{\sigma_{j r}(\bd u)\right\}_{1 \leq j, r \leq p}$ with $\sigma_{j r}(\bd u)=\exp \left(u_1+u_2\right)\phi(u_1 /2+ u_2 /2)^{|j-r|}$.

In Models 1 and 2, while many entries in \(\mb \Sigma(\bd u)\) approach zero as \(p\) becomes sufficiently large, these models are not exactly sparse. Instead, they are designed to evaluate the performance of our method in estimating approximately sparse covariance matrices from non-sparse ones. Tables \ref{table-1} and \ref{table-2} summarize the results of the  MSLs and MFLs for the static methods, FDCMs, modified FDCMs, kernel-based DCM, and modified kernel-based DCM across Models 1 and 2, respectively. Key observations are as follows: 
   (i)  \textbf{Overall Accuracy}:  
    The covariance matrices estimated using our proposed method demonstrate substantial improvements in accuracy compared to static methods across all three values of \(p\) and all four thresholding methods.
     (ii)  \textbf{Model 1 Observations}:  
    In Model 1, the covariance matrix depends solely on \(u_1\). When \(u_1\) is used as the covariate, the kernel-based method can be considered an oracle approach, achieving slightly better accuracy than FDCMs. However, identifying the most informative covariates in practice is challenging. If a non-informative covariate (e.g., \(u_2\)) is mistakenly selected, the kernel-based method's performance deteriorates and becomes worse than that of FDCMs.
    (iii)  \textbf{Model 2 Observations}:  
    In Model 2, both covariates \(u_1\) and \(u_2\) influence the covariance matrix. Even when the kernel-based method uses the informative covariate \(u_1\), its performance still falls short of FDCMs, emphasizing the importance of incorporating multiple covariates. If a non-informative covariate (e.g., \(u_3\)) is used, the kernel-based method performs significantly worse compared to FDCMs.
    (iv) \textbf{Impact of Modifications}:  
    The modified FDCMs and modified kernel-based DCMs exhibit performance similar to their original counterparts, indicating that the modifications do not compromise the accuracy of the estimation methods.

\begin{table}[htbp]
	\centering
	\caption{Average (standard deviation) of MSLs and MFLs for Model 1. $d$ and $p$ represent the dimensions of $\bd U$ and $\bd Y$, respectively.}
	\label{table-1}
	\resizebox{140mm}{!}{
	\begin{tabular}{cccllllll}
		\toprule
		\multirow{2}[4]{*}{} &       &       & \multicolumn{3}{c}{MFL} & \multicolumn{3}{c}{MSL} \\
		\cmidrule{4-9}          & \multicolumn{1}{c}{method} &        & \multicolumn{1}{c}{$p=100$} & \multicolumn{1}{c}{$p=150$} & \multicolumn{1}{c}{$p=200$} & \multicolumn{1}{c}{$p=100$} & \multicolumn{1}{c}{$p=150$} & \multicolumn{1}{c}{$p=200$} \\
		\midrule
		\multirow{29.5}[6]{*}{$d=10$} & \multirow{4}[2]{*}{FDCMs} & Hard & 6.56(1.43) & 8.01(1.73) & 9.29(1.95) & 1.55(0.33) & 1.59(0.41) & 1.56(0.36) \\
		&       & Lasso & 6.44(1.37) & 8.00(1.71) & 9.22(1.96) & 1.52(0.32) & 1.57(0.34) & 1.54(0.34) \\
		&       & Scad & 6.41(1.33) & 7.92(1.66) & 9.21(1.95) & 1.51(0.31) & 1.55(0.34) & 1.54(0.34) \\
		&       & Soft & 6.40(1.33) & 7.92(1.66) & 9.23(1.95) & 1.52(0.31) & 1.55(0.33) & 1.55(0.34) \\
		\cmidrule{2-9}          & \multirow{4}[2]{*}{Modified FDCMs} & Hard & 6.60(1.42) & 8.06(1.75) & 9.32(1.97) & 1.55(0.34) & 1.59(0.41) & 1.57(0.36) \\
		&       & Lasso & 6.44(1.37) & 8.00(1.71) & 9.22(1.95) & 1.52(0.32) & 1.57(0.34) & 1.54(0.34) \\
		&       & Scad & 6.42(1.33) & 7.92(1.67) & 9.21(1.95) & 1.51(0.31) & 1.55(0.34) & 1.54(0.34) \\
		&       & Soft & 6.40(1.33) & 7.92(1.66) & 9.24(1.95) & 1.52(0.31) & 1.55(0.33) & 1.55(0.34) \\
		\cmidrule{2-9}          & \multirow{4}[2]{*}{Static} & Hard & 7.53(0.75) & 9.22(0.89) & 10.63(1.03) & 1.65(0.32) & 1.66(0.33) & 1.65(0.32) \\
		&       & Lasso & 7.26(0.72) & 8.94(0.87) & 10.37(1.02) & 1.62(0.24) & 1.62(0.26) & 1.63(0.25) \\
		&       & Scad & 7.18(0.73) & 8.86(0.89) & 10.30(1.00) & 1.57(0.22) & 1.59(0.27) & 1.59(0.25) \\
		&       & Soft & 7.16(0.73) & 8.84(0.89) & 10.30(1.00) & 1.54(0.20) & 1.57(0.26) & 1.58(0.24) \\
        \cmidrule{2-9}          & \multirow{4}[2]{*}{Kernel$(u_1)$} & Hard & 6.05(1.19) & 7.66(1.34) & 9.00(1.45) & 1.29(0.26) & 1.36(0.24) & 1.37(0.24) \\
		&       & Lasso & 5.33(1.07) & 6.81(1.35) & 8.17(1.51) & 1.22(0.27) & 1.28(0.27) & 1.29(0.28) \\
		&       & Scad & 5.29(1.12) & 6.79(1.33) & 8.07(1.39) & 1.21(0.27) & 1.26(0.27) & 1.27(0.29) \\
		&       & Soft & 5.23(1.10) & 6.76(1.34) & 8.06(1.42) & 1.19(0.28) & 1.24(0.28) & 1.25(0.30) \\
        \cmidrule{2-9}          & \multirow{4}[2]{*}{Kernel$(u_2)$} & Hard & 7.51(0.70) & 9.19(0.84) & 10.60(0.99) & 1.60(0.27) & 1.60(0.28) & 1.56(0.25) \\
		&       & Lasso & 7.25(0.71) & 8.96(0.86) & 10.39(1.01) & 1.62(0.20) & 1.66(0.21) & 1.65(0.18) \\
		&       & Scad & 7.16(0.72) & 8.87(0.87) & 10.30(1.00) & 1.58(0.20) & 1.61(0.21) & 1.61(0.19) \\
		&       & Soft & 7.13(0.70) & 8.87(0.88) & 10.30(1.00) & 1.57(0.19) & 1.59(0.20) & 1.60(0.19) \\
        \cmidrule{2-9}          & \multirow{4}[2]{*}{Modified Kernel$(u_1)$} & Hard & 6.06(1.19) & 7.66(1.34) & 9.00(1.45) & 1.30(0.26) & 1.36(0.24) & 1.37(0.24) \\
		&       & Lasso & 5.33(1.07) & 6.81(1.35) & 8.17(1.51) & 1.22(0.27) & 1.28(0.27) & 1.29(0.28) \\
		&       & Scad & 5.29(1.12) & 6.79(1.33) & 8.07(1.39) & 1.21(0.27) & 1.26(0.27) & 1.27(0.29) \\
		&       & Soft & 5.23(1.10) & 6.76(1.34) & 8.06(1.42) & 1.19(0.28) & 1.24(0.28) & 1.25(0.30) \\
        \cmidrule{2-9}          & \multirow{4}[2]{*}{Modified Kernel$(u_2)$} & Hard & 7.51(0.70) & 9.19(0.84) & 10.60(0.99) & 1.60(0.27) & 1.60(0.28) & 1.56(0.25) \\
		&       & Lasso & 7.25(0.71) & 8.96(0.86) & 10.39(1.01) & 1.62(0.20) & 1.66(0.21) & 1.65(0.18) \\
		&       & Scad & 7.16(0.72) & 8.87(0.87) & 10.30(1.00) & 1.58(0.20) & 1.61(0.21) & 1.61(0.19) \\
		&       & Soft & 7.13(0.70) & 8.87(0.88) & 10.30(1.00) & 1.57(0.19) & 1.59(0.20) & 1.60(0.19) \\
		\midrule
		\multirow{29.5}[6]{*}{$d=20$} & \multirow{4}[2]{*}{FDCMs} & Hard & 6.40(1.33) & 7.86(1.66) & 9.13(1.92) & 1.50(0.33) & 1.55(0.35) & 1.56(0.32) \\
		&       & Lasso & 6.34(1.34) & 7.80(1.67) & 9.04(1.89) & 1.50(0.32) & 1.52(0.33) & 1.52(0.33) \\
		&       & Scad & 6.34(1.34) & 7.83(1.64) & 9.01(1.92) & 1.50(0.32) & 1.52(0.33) & 1.52(0.33) \\
		&       & Soft & 6.33(1.35) & 7.79(1.66) & 9.03(1.90) & 1.50(0.32) & 1.52(0.33) & 1.52(0.33) \\
		\cmidrule{2-9}          & \multirow{4}[2]{*}{Modified FDCMs} & Hard & 6.40(1.33) & 7.87(1.66) & 9.21(1.89) & 1.51(0.33) & 1.56(0.37) & 1.56(0.32) \\
		&       & Lasso & 6.34(1.34) & 7.80(1.67) & 9.04(1.89) & 1.50(0.32) & 1.52(0.33) & 1.52(0.33) \\
		&       & Scad & 6.35(1.34) & 7.85(1.63) & 9.01(1.92) & 1.50(0.32) & 1.52(0.33) & 1.52(0.33) \\
		&       & Soft & 6.33(1.35) & 7.79(1.65) & 9.03(1.90) & 1.50(0.32) & 1.52(0.33) & 1.52(0.33) \\
		\cmidrule{2-9}          & \multirow{4}[2]{*}{Static} & Hard & 7.53(0.75) & 9.22(0.89) & 10.63(1.03) & 1.65(0.32) & 1.66(0.33) & 1.65(0.32) \\
		&       & Lasso & 7.26(0.72) & 8.94(0.87) & 10.37(1.02) & 1.62(0.24) & 1.62(0.26) & 1.63(0.25) \\
		&       & Scad & 7.18(0.73) & 8.86(0.89) & 10.30(1.00) & 1.57(0.22) & 1.59(0.27) & 1.59(0.25) \\
		&       & Soft & 7.16(0.73) & 8.84(0.89) & 10.30(1.00) & 1.54(0.20) & 1.57(0.26) & 1.58(0.24) \\
        \cmidrule{2-9}          & \multirow{4}[2]{*}{Kernel$(u_1)$} & Hard & 6.05(1.19) & 7.67(1.33) & 9.00(1.46) & 1.29(0.26) & 1.36(0.24) & 1.37(0.24) \\
		&       & Lasso & 5.33(1.07) & 6.82(1.34) & 8.16(1.51) & 1.22(0.27) & 1.28(0.27) & 1.29(0.28) \\
		&       & Scad & 5.29(1.12) & 6.79(1.33) & 8.08(1.40) & 1.21(0.27) & 1.26(0.27) & 1.27(0.29) \\
		&       & Soft & 5.23(1.10) & 6.74(1.34) & 8.05(1.42) & 1.19(0.28) & 1.25(0.28) & 1.25(0.30) \\
        \cmidrule{2-9}          & \multirow{4}[2]{*}{Kernel$(u_2)$} & Hard & 7.49(0.70) & 9.19(0.84) & 10.60(0.99) & 1.60(0.26) & 1.59(0.26) & 1.55(0.25) \\
		&       & Lasso & 7.24(0.71) & 8.96(0.86) & 10.39(1.01) & 1.62(0.20) & 1.66(0.21) & 1.65(0.19) \\
		&       & Scad & 7.16(0.71) & 8.87(0.88) & 10.29(1.00) & 1.58(0.20) & 1.61(0.21) & 1.61(0.19) \\
		&       & Soft & 7.14(0.70) & 8.87(0.89) & 10.29(1.00) & 1.57(0.20) & 1.59(0.20) & 1.60(0.19) \\
        \cmidrule{2-9}          & \multirow{4}[2]{*}{Modified Kernel$(u_1)$} & Hard & 6.06(1.19) & 7.67(1.33) & 9.00(1.46) & 1.30(0.26) & 1.36(0.24) & 1.37(0.24) \\
		&       & Lasso & 5.33(1.07) & 6.82(1.34) & 8.16(1.51) & 1.22(0.27) & 1.28(0.27) & 1.29(0.28) \\
		&       & Scad & 5.29(1.12) & 6.79(1.33) & 8.08(1.40) & 1.21(0.27) & 1.26(0.27) & 1.27(0.29) \\
		&       & Soft & 5.23(1.10) & 6.74(1.34) & 8.05(1.42) & 1.19(0.28) & 1.25(0.28) & 1.25(0.30) \\
        \cmidrule{2-9}          & \multirow{4}[2]{*}{Modified Kernel$(u_2)$} & Hard & 7.49(0.70) & 9.19(0.84) & 10.60(0.99) & 1.60(0.26) & 1.59(0.26) & 1.55(0.25) \\
		&       & Lasso & 7.24(0.71) & 8.96(0.86) & 10.39(1.01) & 1.62(0.20) & 1.66(0.21) & 1.65(0.19) \\
		&       & Scad & 7.16(0.71) & 8.87(0.88) & 10.29(1.00) & 1.58(0.20) & 1.61(0.21) & 1.61(0.19) \\
		&       & Soft & 7.14(0.70) & 8.87(0.89) & 10.29(1.00) & 1.57(0.20) & 1.59(0.20) & 1.60(0.19) \\
		\bottomrule
	\end{tabular}%
}
\end{table}%

\begin{table}[htbp]
	\centering
	\caption{Average (standard deviation) of MSLs and MFLs for Model 2. $d$ and $p$ represent  the dimensions of $\bd U$ and $\bd Y$, respectively.}
	\label{table-2}
	\resizebox{140mm}{!}{%
	\begin{tabular}{cccllllll}
		\toprule
		\multirow{2}[4]{*}{} &       &       & \multicolumn{3}{c}{MFL} & \multicolumn{3}{c}{MSL} \\
		\cmidrule{4-9}          & \multicolumn{1}{c}{method} &        & \multicolumn{1}{c}{$p=100$} & \multicolumn{1}{c}{$p=150$} & \multicolumn{1}{c}{$p=200$} & \multicolumn{1}{c}{$p=100$} & \multicolumn{1}{c}{$p=150$} & \multicolumn{1}{c}{$p=200$} \\
		\midrule
		\multirow{29.5}[6]{*}{$d=10$} & \multirow{4}[2]{*}{FDCMs} & Hard & 8.50(2.16) & 10.73(2.85) & 12.42(3.29) & 2.08(0.63) & 2.30(0.75) & 2.35(0.80) \\
		&       & Lasso & 8.38(2.14) & 10.34(2.52) & 11.92(2.97) & 1.97(0.56) & 2.06(0.60) & 2.10(0.61) \\
		&       & Scad & 8.36(2.04) & 10.31(2.56) & 11.88(2.93) & 1.93(0.53) & 2.03(0.58) & 2.06(0.58) \\
		&       & Soft & 8.35(2.04) & 10.32(2.56) & 11.94(2.89) & 1.94(0.52) & 2.06(0.57) & 2.09(0.58) \\
		\cmidrule{2-9}          & \multirow{4}[2]{*}{Modified FDCMs} & Hard & 8.63(2.19) & 11.12(3.36) & 13.07(3.91) & 2.10(0.64) & 2.33(0.80) & 2.35(0.80) \\
		&       & Lasso & 8.38(2.14) & 10.36(2.52) & 11.95(2.98) & 1.97(0.56) & 2.06(0.60) & 2.10(0.62) \\
		&       & Scad & 8.37(2.05) & 10.33(2.57) & 11.90(2.93) & 1.93(0.53) & 2.03(0.58) & 2.06(0.58) \\
		&       & Soft & 8.35(2.04) & 10.32(2.56) & 11.95(2.89) & 1.94(0.52) & 2.06(0.57) & 2.09(0.58) \\
		\cmidrule{2-9}          & \multirow{4}[2]{*}{Static} & Hard & 9.90(1.25) & 12.13(1.47) & 14.00(1.70) & 2.26(0.52) & 2.35(0.44) & 2.36(0.49) \\
		&       & Lasso & 9.68(1.20) & 11.90(1.40) & 13.78(1.63) & 2.18(0.39) & 2.17(0.33) & 2.16(0.35) \\
		&       & Scad & 9.62(1.19) & 11.81(1.43) & 13.70(1.64) & 2.11(0.36) & 2.15(0.32) & 2.15(0.34) \\
		&       & Soft & 9.58(1.18) & 11.80(1.44) & 13.70(1.64) & 2.14(0.37) & 2.15(0.32) & 2.14(0.33) \\
        \cmidrule{2-9}          & \multirow{4}[2]{*}{Kernel$(u_1)$} & Hard & 8.99(1.74) & 11.03(2.01) & 12.73(1.99) & 1.93(0.37) & 1.99(0.42) & 1.94(0.35) \\
		&       & Lasso & 8.69(1.73) & 10.77(1.97) & 12.54(1.95) & 1.99(0.37) & 2.03(0.38) & 2.00(0.32) \\
		&       & Scad & 8.59(1.73) & 10.65(1.97) & 12.42(1.94) & 1.98(0.38) & 1.99(0.38) & 1.96(0.33) \\
		&       & Soft & 8.57(1.73) & 10.62(1.98) & 12.40(1.94) & 1.97(0.38) & 1.97(0.39) & 1.95(0.33) \\
        \cmidrule{2-9}          & \multirow{4}[2]{*}{Kernel$(u_3)$} & Hard & 9.86(1.27) & 12.09(1.56) & 13.96(1.78) & 1.95(0.31) & 2.03(0.40) & 2.01(0.29) \\
		&       & Lasso & 9.76(1.32) & 11.98(1.61) & 13.88(1.83) & 2.16(0.32) & 2.18(0.36) & 2.14(0.32) \\
		&       & Scad & 9.65(1.35) & 11.89(1.62) & 13.80(1.82) & 2.13(0.34) & 2.18(0.35) & 2.15(0.31) \\
		&       & Soft & 9.64(1.33) & 11.88(1.63) & 13.79(1.83) & 2.15(0.33) & 2.17(0.35) & 2.16(0.31) \\
        \cmidrule{2-9}          & \multirow{4}[2]{*}{Modified Kernel$(u_1)$} & Hard & 8.99(1.74) & 11.03(2.01) & 12.73(1.99) & 1.93(0.37) & 1.99(0.42) & 1.94(0.35) \\
		&       & Lasso & 8.69(1.73) & 10.77(1.97) & 12.54(1.95) & 1.99(0.37) & 2.03(0.38) &2.00(0.32)\\
		&       & Scad & 8.59(1.73) & 10.65(1.97) & 12.42(1.94) & 1.98(0.38) & 1.99(0.38) & 1.96(0.33) \\
		&       & Soft & 8.57(1.73) & 10.62(1.98) & 12.40(1.94) & 1.97(0.38) & 1.97(0.39) & 1.95(0.33) \\
        \cmidrule{2-9}          & \multirow{4}[2]{*}{Modified Kernel$(u_3)$} & Hard & 9.86(1.27) & 12.09(1.56) & 13.96(1.78) & 1.95(0.31) & 2.03(0.40) & 2.01(0.29) \\
		&       & Lasso & 9.76(1.32) & 11.98(1.61) & 13.88(1.83) & 2.16(0.32) & 2.18(0.36) & 2.14(0.32) \\
		&       & Scad & 9.65(1.35) & 11.89(1.62) & 13.80(1.82) & 2.13(0.34) & 2.18(0.35) & 2.15(0.31) \\
		&       & Soft & 9.64(1.33) & 11.88(1.63) & 13.79(1.83) & 2.15(0.33) & 2.17(0.35) & 2.16(0.31) \\
		\midrule
		\multirow{29.5}[6]{*}{$d=20$} & \multirow{4}[2]{*}{FDCMs} & Hard & 8.41(1.80) & 10.73(2.85) & 12.25(2.83) & 1.97(0.52) & 2.30(0.75) & 2.15(0.61) \\
		&       & Lasso & 8.35(1.71) & 10.34(2.52) & 11.91(2.48) & 1.92(0.49) & 2.06(0.60) & 2.04(0.50) \\
		&       & Scad & 8.31(1.67) & 10.31(2.56) & 11.88(2.42) & 1.87(0.48) & 2.03(0.58) & 2.02(0.53) \\
		&       & Soft & 8.31(1.66) & 10.32(2.56) & 11.87(2.42) & 1.87(0.47) & 2.06(0.57) & 2.02(0.54) \\
		\cmidrule{2-9}          & \multirow{4}[2]{*}{Modified FDCMs} & Hard & 8.50(1.86) & 11.12(3.36) & 12.46(2.97) & 1.98(0.52) & 2.33(0.80) & 2.16(0.62) \\
		&       & Lasso & 8.35(1.71) & 10.36(2.52) & 11.91(2.48) & 1.92(0.49) & 2.06(0.60) & 2.04(0.50) \\
		&       & Scad & 8.31(1.67) & 10.33(2.57) & 11.96(2.55) & 1.87(0.48) & 2.03(0.58) & 2.02(0.53) \\
		&       & Soft & 8.31(1.66) & 10.32(2.56) & 11.88(2.42) & 1.87(0.47) & 2.06(0.57) & 2.02(0.54) \\
		\cmidrule{2-9}          & \multirow{4}[2]{*}{Static} & Hard & 9.90(1.25) & 12.13(1.47) & 14.00(1.70) & 2.26(0.52) & 2.35(0.44) & 2.36(0.49) \\
		&       & Lasso & 9.68(1.20) & 11.90(1.40) & 13.78(1.63) & 2.18(0.39) & 2.17(0.33) & 2.16(0.35) \\
		&       & Scad & 9.62(1.19) & 11.81(1.43) & 13.70(1.64) & 2.11(0.36) & 2.15(0.32) & 2.15(0.34) \\
		&       & Soft & 9.58(1.18) & 11.80(1.44) & 13.70(1.64) & 2.14(0.37) & 2.15(0.32) & 2.14(0.33) \\ 
        \cmidrule{2-9}          & \multirow{4}[2]{*}{Kernel$(u_1)$} & Hard & 8.98(1.74) & 11.03(2.01) & 12.74(2.00) & 1.94(0.37) & 1.99(0.42) & 1.96(0.37) \\
		&       & Lasso & 8.68(1.73) & 10.78(1.96) & 12.55(1.94) & 1.99(0.37) & 2.03(0.38) & 2.00(0.33) \\
		&       & Scad & 8.59(1.73) & 10.65(1.97) & 12.42(1.94) & 1.97(0.38) & 2.00(0.38) & 1.96(0.33) \\
		&       & Soft & 8.55(1.73) & 10.63(1.97) & 12.41(1.94) & 1.96(0.38) & 1.97(0.39) & 1.95(0.33) \\
        \cmidrule{2-9}          & \multirow{4}[2]{*}{Kernel$(u_3)$} & Hard & 9.87(1.27) & 12.08(1.56) & 13.96(1.78) & 1.97(0.33) & 2.02(0.39) & 2.01(0.30) \\
		&       & Lasso & 9.75(1.33) & 11.98(1.61) & 13.89(1.82) & 2.15(0.31) & 2.18(0.36) & 2.15(0.32) \\
		&       & Scad & 9.65(1.34) & 11.89(1.61) & 13.80(1.83) & 2.13(0.33) & 2.18(0.36) & 2.15(0.31) \\
		&       & Soft & 9.64(1.34) & 11.88(1.62) & 13.80(1.83) & 2.14(0.33) & 2.17(0.35) & 2.16(0.31) \\
        \cmidrule{2-9}          & \multirow{4}[2]{*}{Modified Kernel$(u_1)$} & Hard & 8.98(1.74) & 11.03(2.01) & 12.74(2.00) & 1.94(0.37) & 1.99(0.42) & 1.96(0.37) \\
		&       & Lasso & 8.68(1.73) & 10.78(1.96) & 12.55(1.94) & 1.99(0.37) & 2.03(0.38) & 2.00(0.33) \\
		&       & Scad & 8.59(1.73) & 10.65(1.97) & 12.42(1.94) & 1.97(0.38) & 2.00(0.38) & 1.96(0.33) \\
		&       & Soft & 8.55(1.73) & 10.63(1.97) & 12.41(1.94) & 1.96(0.38) & 1.97(0.39) & 1.95(0.33) \\
        \cmidrule{2-9}          & \multirow{4}[2]{*}{Modified Kernel$(u_3)$} & Hard & 9.87(1.27) & 12.08(1.56) & 13.96(1.78) & 1.97(0.33) & 2.02(0.39) & 2.01(0.30) \\
		&       & Lasso & 9.75(1.33) & 11.98(1.61) & 13.89(1.82) & 2.15(0.31) & 2.18(0.36) & 2.15(0.32) \\
		&       & Scad & 9.65(1.34) & 11.89(1.61) & 13.80(1.83) & 2.13(0.33) & 2.18(0.36) & 2.15(0.31) \\
		&       & Soft & 9.64(1.34) & 11.88(1.62) & 13.80(1.83) & 2.14(0.33) & 2.17(0.35) & 2.16(0.31) \\
		\bottomrule
	\end{tabular}%
}%
\end{table}%

\begin{table}[htb]
	\centering
\caption{Average (standard deviation) of MTPRs and MFPRs for Model 3.  $d$ and $p$ represent the dimensions of $\bd U$ and $\bd Y$, respectively.}
\label{table-3}
\resizebox{140mm}{!}{%
	\begin{tabular}{cccllllll}
		\toprule
		\multirow{2}[4]{*}{} &       &       & \multicolumn{3}{c}{MTPR} & \multicolumn{3}{c}{MFPR} \\
		\cmidrule{4-9}          & method &       & \multicolumn{1}{c}{$p=100$} & \multicolumn{1}{c}{$p=150$} & \multicolumn{1}{c}{$p=200$} & \multicolumn{1}{c}{$p=100$} & \multicolumn{1}{c}{$p=150$} & \multicolumn{1}{c}{$p=200$} \\
		\midrule
		\multirow{16.5}[4]{*}{$d=10$} & \multirow{4}[2]{*}{FDCMs} & Hard & 0.36(0.03) & 0.36(0.02) & 0.35(0.02) & 0.00(0.00) & 0.00(0.00) & 0.00(0.00) \\
		&       & Lasso & 0.38(0.04) & 0.38(0.04) & 0.37(0.03) & 0.00(0.00) & 0.00(0.00) & 0.00(0.00) \\
		&       & Scad & 0.40(0.05) & 0.39(0.04) & 0.38(0.04) & 0.00(0.00) & 0.00(0.00) & 0.00(0.00) \\
		&       & Soft & 0.41(0.06) & 0.40(0.04) & 0.39(0.05) & 0.00(0.00) & 0.00(0.00) & 0.00(0.00) \\
		\cmidrule{2-9}          & \multirow{4}[2]{*}{Static} & Hard & 0.33(0.03) & 0.33(0.03) & 0.33(0.03) & 0.00(0.00) & 0.00(0.00) & 0.00(0.00) \\
		&       & Lasso & 0.35(0.03) & 0.34(0.03) & 0.33(0.03) & 0.00(0.00) & 0.00(0.00) & 0.00(0.00) \\
		&       & Scad & 0.39(0.03) & 0.35(0.03) & 0.33(0.03) & 0.00(0.00) & 0.00(0.00) & 0.00(0.00) \\
		&       & Soft & 0.39(0.03) & 0.35(0.03) & 0.33(0.03) & 0.00(0.00) & 0.00(0.00) & 0.00(0.00) \\
        \cmidrule{2-9}          & \multirow{4}[2]{*}{Kernel$(u_1)$} & Hard & 0.34(0.02) & 0.33(0.02) & 0.33(0.02) & 0.00(0.00) & 0.00(0.00) & 0.00(0.00) \\
		&       & Lasso & 0.51(0.10) & 0.47(0.09) &0.44(0.08) & 0.00(0.00) & 0.00(0.00) & 0.00(0.00) \\
		&       & Scad & 0.61(0.12) & 0.58(0.11) & 0.53(0.11) & 0.00(0.00) & 0.00(0.00) & 0.00(0.00) \\
		&       & Soft & 0.63(0.13) & 0.59(0.11) & 0.54(0.11) & 0.00(0.00) & 0.00(0.00) & 0.00(0.00) \\
        \cmidrule{2-9}          & \multirow{4}[2]{*}{Kernel$(u_2)$} & Hard & 0.33(0.03) & 0.33(0.03) & 0.33(0.03) & 0.00(0.00) & 0.00(0.00) & 0.00(0.00) \\
		&       & Lasso & 0.33(0.03) & 0.33(0.03) & 0.33(0.03) & 0.00(0.00) & 0.00(0.00) & 0.00(0.00) \\
		&       & Scad & 0.35(0.05) & 0.34(0.03) & 0.33(0.03) & 0.00(0.00) & 0.00(0.00) & 0.00(0.00) \\
		&       & Soft & 0.35(0.05) & 0.34(0.03) & 0.33(0.03) & 0.00(0.00) & 0.00(0.00) & 0.00(0.00) \\
		\midrule
		\multirow{16.5}[4]{*}{$d=20$} & \multirow{4}[2]{*}{FDCMs} & Hard & 0.35(0.02) & 0.35(0.02) & 0.35(0.02) & 0.00(0.00) & 0.00(0.00) & 0.00(0.00) \\
		&       & Lasso & 0.38(0.04) & 0.37(0.03) & 0.37(0.03) & 0.00(0.00) & 0.00(0.00) & 0.00(0.00) \\
		&       & Scad & 0.40(0.04) & 0.38(0.04) & 0.38(0.03) & 0.00(0.00) & 0.00(0.00) & 0.00(0.00) \\
		&       & Soft & 0.40(0.05) & 0.39(0.04) & 0.38(0.04) & 0.00(0.00) & 0.00(0.00) & 0.00(0.00) \\
		\cmidrule{2-9}          & \multirow{4}[2]{*}{Static} & Hard & 0.33(0.03) & 0.33(0.03) & 0.33(0.03) & 0.00(0.00) & 0.00(0.00) & 0.00(0.00) \\
		&       & Lasso & 0.35(0.03) & 0.34(0.03) & 0.33(0.03) & 0.00(0.00) & 0.00(0.00) & 0.00(0.00) \\
		&       & Scad & 0.39(0.03) & 0.35(0.03) & 0.33(0.03) & 0.00(0.01) & 0.00(0.00) & 0.00(0.00) \\
		&       & Soft & 0.39(0.03) & 0.35(0.03) & 0.33(0.03) & 0.00(0.01) & 0.00(0.00) & 0.00(0.00) \\
        \cmidrule{2-9}          & \multirow{4}[2]{*}{Kernel$(u_1)$} & Hard & 0.34(0.02) & 0.33(0.02) & 0.33(0.02) & 0.00(0.00) & 0.00(0.00) & 0.00(0.00) \\
		&       & Lasso & 0.50(0.10) & 0.47(0.09) & 0.44(0.08) & 0.00(0.00) & 0.00(0.00) & 0.00(0.00) \\
		&       & Scad & 0.61(0.13) & 0.58(0.11) & 0.53(0.11) & 0.00(0.00) & 0.00(0.00) & 0.00(0.00) \\
		&       & Soft & 0.63(0.13) & 0.59(0.11) & 0.53(0.11) & 0.01(0.00) & 0.00(0.00) & 0.00(0.00) \\
        \cmidrule{2-9}          & \multirow{4}[2]{*}{Kernel$(u_2)$} & Hard & 0.33(0.03) & 0.33(0.03) & 0.33(0.03) & 0.00(0.00) & 0.00(0.00) & 0.00(0.00) \\
		&       & Lasso & 0.33(0.03) & 0.33(0.03) & 0.33(0.03) & 0.00(0.00) & 0.00(0.00) & 0.00(0.00) \\
		&       & Scad & 0.35(0.04) & 0.34(0.03) & 0.33(0.03) & 0.00(0.00) & 0.00(0.00) & 0.00(0.00) \\
		&       & Soft & 0.35(0.05) & 0.34(0.03) & 0.33(0.03) & 0.00(0.00) & 0.00(0.00) & 0.00(0.00) \\
		\bottomrule
	\end{tabular}%
}%
\end{table}%

\subsection*{Varying-Sparsity Covariance Models:}
\noindent \textbf{Model 3:} $\bd Y \sim N\left(\mathbf{0}, \mb \Sigma(\bd u)\right),$ where $\mb \Sigma(\bd u)=\left\{\sigma_{j r}(\bd u)\right\}_{1 \leq j, r \leq p}$ with
\begin{equation*}
    \begin{aligned}
        \sigma_{j r}(\bd u)&=\exp \left(2u_1\right)\big{[}0.5 \exp \left\{-\frac{\left(u_1-0.25\right)^2}{0.75^2-\left(u_1-0.25\right)^2}\right\} \mathbb{I}\left\{-0.5 \leq u_1 \leq 1\right\} \mathbb{I}\{|{j}-{r}|=1\} \\
	&~~~~~~~~+0.4 \exp \left\{-\frac{\left(u_1-0.65\right)^2}{0.35^2-\left(u_1-0.65\right)^2}\right\} \mathbb{I}\left\{0.3 \leq u_1 \leq 1\right\} \mathbb{I}\{|{j}-{r}|=2\}+\mathbb{I}\{{j}={r}\}\big{]}.
    \end{aligned}
\end{equation*} 

\noindent \textbf{Model 4:} $\bd Y \sim N\left(\mathbf{0}, \mb \Sigma(\bd u)\right),$ where $\mb \Sigma(\bd u)=\left\{\sigma_{j r}(\bd u)\right\}_{1 \leq j, r \leq p}$ with $\sigma_{j r}(\bd u)=0.5\zeta_{j r}(u_1,u_2)+0.5\zeta_{j r}(u_2,u_1)$ and
\begin{equation*}
    \begin{aligned}
        \zeta_{j r}(u_1,u_2)=&\exp (2u_1)\Big[0.5 \exp \left\{-\frac{(u_1-0.25)^2}{0.75^2-(u_1-0.25)^2}\right\} \mathbb{I}\{-0.5 \leq u_1 \leq 1,-0.5 \leq u_2 \leq 1\}  \\
	& \mathbb{I}\{|j-r|=1\}+0.4 \exp \left\{-\frac{(u_1-0.65)^2}{0.35^2-(u_1-0.65)^2}\right\}  
	\mathbb{I}\{0.3 \leq u_1 \leq 1,0.3 \leq u_2 \leq 1\}  \\
	& \mathbb{I}\{|j-r|=2\}+\mathbb{I}\{j=r\}\Big].
    \end{aligned}
\end{equation*} 

In Models 3 and 4, the sparsity of the dynamic covariance model depends on the value of $\bd u$.
The performance of recovering the varying sparsity is evaluated using the medians of the True Positive Rates (TPR) and False Positive Rates (FPR) calculated at the 30 test points, defined as follows:
\begin{equation*}
    \begin{aligned}
        & \text { Median TPR }=\operatorname{median}\left\{\text{TPR}({\bd u_k}),k\in [30]\right\}, \\
        & \text { Median FPR}=\operatorname{median}\left\{\text{FPR}({\bd u_k}),k\in [30]\right\}, 
    \end{aligned}
\end{equation*} where 
\begin{equation*}
    \begin{aligned}
        & \text{TPR}(\bd u_k)=\frac{|\left\{(j, r): s_{\lambda_n(\bd u_k)}\left(\widehat{\sigma}_{j r}(\bd u_k)\right) \neq 0 \text { and } \sigma_{j r}(\bd u_k) \neq 0\right\}|}{|\left\{(j, r): \sigma_{j r}(\bd u_k) \neq 0\right\}|}, \\
	&  \text{FPR}(\bd u_k) =\frac{|\left\{(j, r): s_{\lambda_n(\bd u_k)}\left(\widehat{\sigma}_{j r}(\bd u_k)\right) \neq 0 \text { and } \sigma_{j r}(\bd u_k)=0\right\}|}{|\left\{(j, r): \sigma_{j r}(\bd u_k)=0\right\}|}.
    \end{aligned}
\end{equation*} For simplicity, we refer to the median TPR as MTPR and the median FPR as MFPR.  We  compute the average and standard deviation of the MTPRs and MFPR  over 50 replications,  respectively. Since the sparsity remains unchanged in the modified FDCMs and kernel-based DCM compared to their  pre-modification versions, we do not report MTPR and MFPR for the two methods. Furthermore,  we calculate both MSLs and MFLs to assess the  accuracy of the covariance matrices estimators.

Tables S.1 and S.2 (Supplementary Materials S.5) present the results of the MSLs and MFLs for the static method, FDCMs, modified FDCMs, kernel-based DCM, and modified kernel-based DCM in Models 3 and 4, respectively. In addition, Tables \ref{table-3} and S.3 (Supplementary Materials S.5) provide the MTPRs and MFPRs.  
In Model 3, the covariance matrix is determined solely by \(u_1\), while in Model 4, both \(u_1\) and \(u_2\) simultaneously influence the covariance matrix. Based on the results, we make the following observations: 
(i) \textbf{Consistency Across Models}:  
    The conclusions regarding MSLs and MFLs from Models 1 and 2 remain consistent in Models 3 and 4, respectively.
(ii) \textbf{Higher True Positive Rates}:  
    Our FDCM consistently achieves higher MTPRs compared to the static method, with particularly strong performance observed under the soft and SCAD thresholding rules.
(iii) \textbf{Performance in Model 3}:  
    In Model 3, the kernel-based DCM method that incorporates the significant covariate \(u_1\) outperforms FDCM in terms of MTPR. However, when the kernel-based DCM method uses the non-significant covariate \(u_2\), its MTPR is lower than that of FDCM.
    (iv) \textbf{Performance in Model 4}:  
    In Model 4, for all three values of \(p\) and across all four thresholding methods, our FDCM consistently outperforms the kernel-based DCM methods. This holds true regardless of whether the kernel-based DCM uses the significant covariate \(u_1\) or the non-significant covariate \(u_3\).

\section{Real Data Analysis}
\label{Real Data} 

In this section, we focus on constructing a global minimum-variance portfolio, which aims to minimize risk by optimizing the weights of a given set of assets to reduce the portfolio’s variance \citep{reh2023predicting}. 
We use Datastream to collect daily returns (in percentages) of stocks listed on the New York Stock Exchange over the period from January 2, 2014, to December 29, 2023. From this dataset, we select 200 stocks with the largest market capitalizations at the end of the period, ensuring that all selected stocks maintained a complete trading history throughout the observation window. 
As conditioning covariates, we incorporate the one-day-lagged returns of the Fama–French five factors \citep{fama2015five}, which are widely recognized in academia and financial practice for their impact on stock returns. These factors include: 
 market risk (Mkt-RF), size (SMB), value (HML), profitability (RMW), and investment (CMA). 
 The Fama–French factor data are obtained from Kenneth French’s data library (\url{https://mba.tuck.dartmouth.edu/pages/faculty/ken.french/data_library.html}). 
The out-of-sample period for evaluation spans from May 28, 2014, to December 29, 2023, providing an extensive window for assessing portfolio performance.

Let $\bd{U}_i \in \mathbb{R}^5$ represent the Fama-French five-factor return vector on day $i$, and let $\bd Y_{i}\in \mathbb{R}^{200}$ denote the returns of 200 stocks on the following day.
Given  $\bd U_i$, the covariance matrix of $\bd Y_{i}$ can be modeled as $\mb \Sigma(\bd U_i)=\mbox{Cov}(\bd Y_i|\bd U_i)\in \mathbb{R}^{200\times 200}$.
Then, for  day $i+1$, the global minimum-variance portfolio is obtained by solving:
\begin{equation*}
    \begin{aligned}
        \argmin_{\bd w} \left\{ \bd w^{\top}\mb \Sigma(\bd U_i)\bd w\right\}, \text{ s.t. } \bd w^{\top}\bd 1_{p}=1,
    \end{aligned}
\end{equation*} with $p=200$ \citep{engle2017large, chen2019new}.  The analytical solution to this problem can be written as: $\bd w_i^* = {\mb \Sigma^{-1}(\bd U_i)\bd 1_{p}}/\{\bd 1_{p}^{\top} \mb \Sigma^{-1}(\bd U_i)\bd 1_{p}\}.$ In our study, we allow  short selling, i.e.,  elements of  $\bd w_t^*$ can be negative, and we assume that there are no transaction costs. 
Once we obtain $\widehat{\mb \Sigma}(\cdot)$,   $\bd w_i^*$ can be estimated as $\widehat{\bd w}_i={\widehat{\mb \Sigma}^{-1}(\bd U_i)\bd 1_{p}}/\{\bd 1_{p}^{\top} \widehat{\mb \Sigma}^{-1}(\bd U_i)\bd 1_{p}\}.$
Then, the  return for day $i+1$ under the minimum-variance portfolio \citep{kempf2006estimating} can be estimated as 
\begin{equation}\label{Ri}
   \widehat{ R}_{i+1}=\widehat{\bd w}^{\top}_i\bd Y_{i}.
\end{equation} 

\subsection{The dynamic covariance  models}

\begin{figure}[htbp]
    \centering
    \begin{minipage}[t]{\linewidth}
        \centering
        \includegraphics[width=160mm]{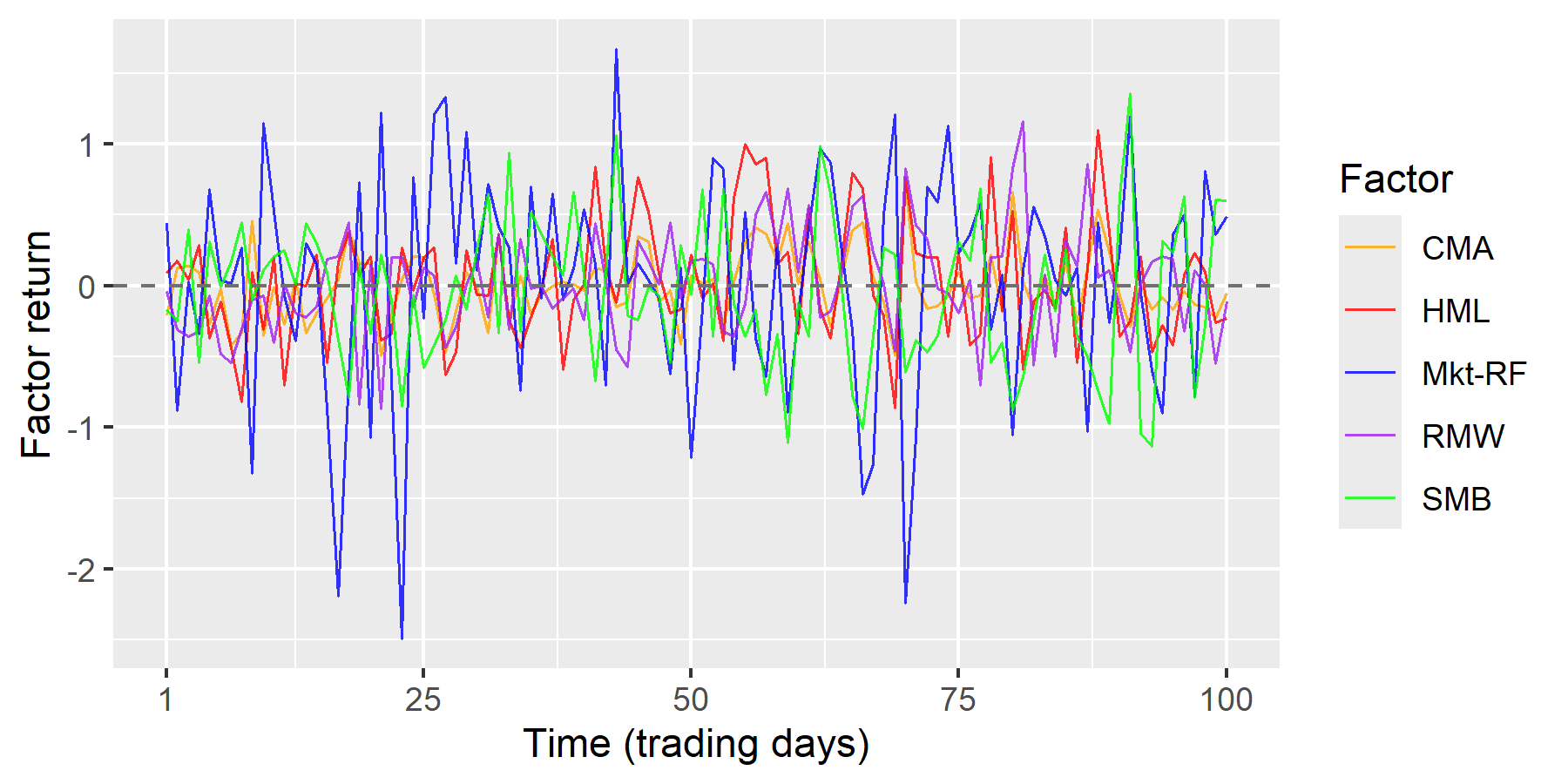}
    \caption{Returns on the Fama-French five factors as functions of time. The horizontal axis from 1 to 100 corresponds to the first 100 trading days from January 2, 2014, to December 29, 2023, arranged in chronological order.}
    \label{Time-dependent}
    \vspace{10mm}
    \end{minipage}
    \begin{minipage}[t]{\linewidth}
        \centering
        \includegraphics[width=160mm]{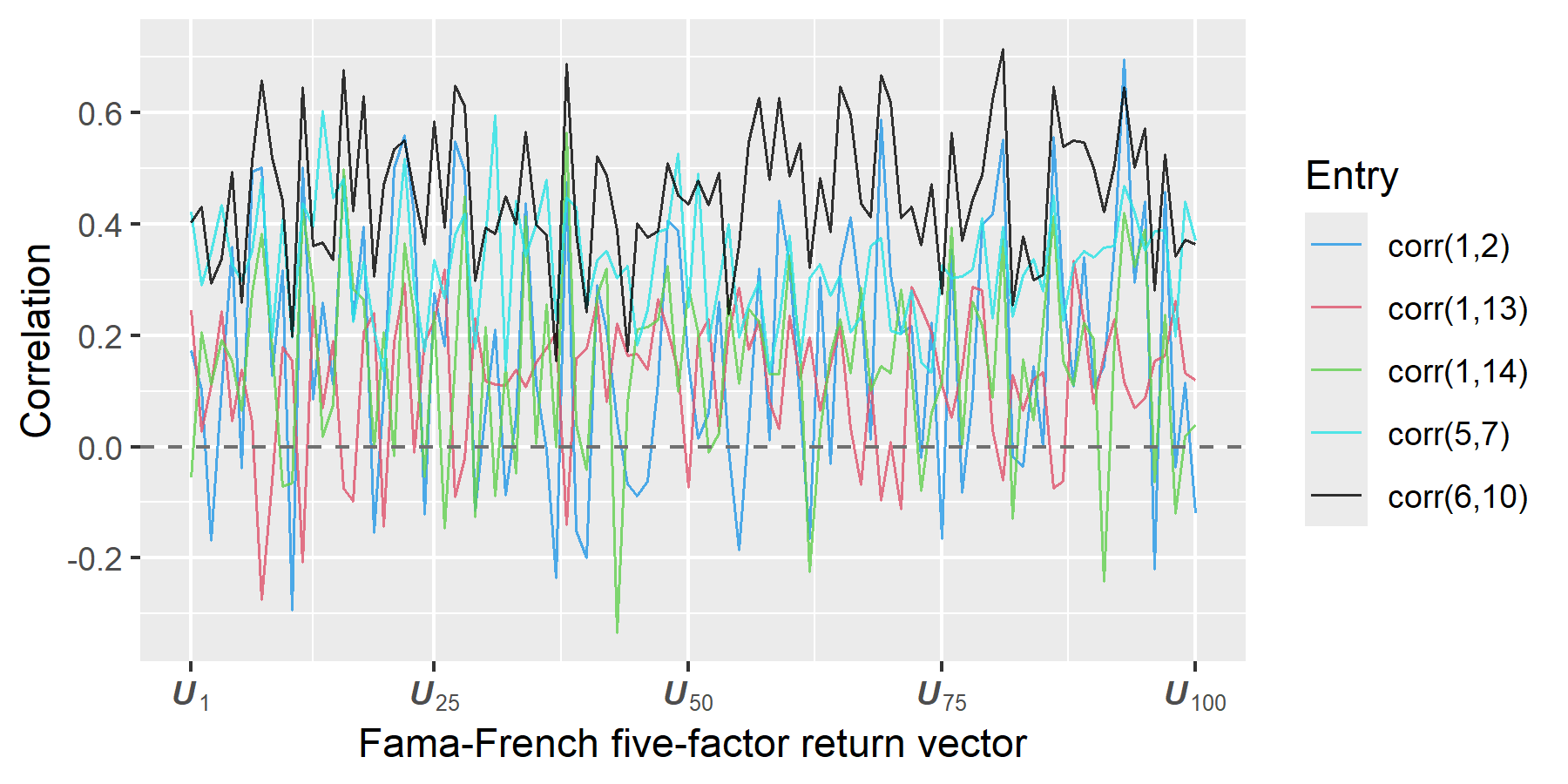}
    \caption{Selected entries of the correlation matrix as functions of Fama-French five-factor return vector using (\ref{covariancehat}). The horizontal axis from $\bd U_1$ to $\bd U_{100}$ corresponds to returns on the Fama-French five-factor return vector at the first 100 trading days from January 2, 2014, to December
29, 2023 and $\text{corr}(j,r)$ denotes the $(j,r)$-th element of the sample correlation matrix. \label{fig-1}}
    \end{minipage}
\end{figure}

We provide a detailed rationale for employing dynamic covariance models. To illustrate this, we consider the first 100 trading days within the period from January 2, 2014, to December 29, 2023, labeling these days sequentially from 1 to 100 in chronological order. 
First, Figure \ref{Time-dependent} depicts the temporal dynamics of returns for the Fama–French five factors. The results reveal significant fluctuations in factor returns over time, including multiple instances of sign reversals, highlighting the inherent time-dependent nature of these returns. 
Second, using data from these 100 days, we train a random forest model and compute the sample dynamic covariance matrix \(\widehat{\mb \Sigma}(\bd U_i)\) as defined in (\ref{covariancehat}) at the Fama–French five-factor return vector \(\bd U_i\), for \(i = 1, \ldots, 100\). From these estimates, we calculate the corresponding sample correlation matrices and plot selected entries in Figure \ref{fig-1}. The results demonstrate that the correlations between different stocks vary with changes in the Fama–French five-factor return vector. For example:
\begin{itemize}
    \item Correlation entries such as \((1,2)\), \((1,13)\), and \((1,14)\) exhibit multiple sign reversals as functions of the Fama–French five-factor return vector.
    \item Correlation magnitudes for entries \((5,7)\) and \((6,10)\) show significant variation over time.
\end{itemize} 
These findings indicate that stock correlations are dynamic and influenced by changes in the Fama–French five-factor return vector. This motivates us to conclude that the proposed dynamic covariance models are better suited for capturing these time-dependent relationships.


For the sparse matrix estimation, we report results exclusively using the soft thresholding rule, as simulation studies have shown that this approach effectively captures the true sparsity of covariance matrices. Our proposed covariance matrix estimators are computed at three specific points: 
$
\bd U_1 = (0.44, -0.17, 0.09, -0.04, -0.21)^T,$ $  
\bd U_{50} = (-1.21, -0.06, 0.22,\\ 0.18, 0.07)^T,$ and $ 
\bd U_{100} = (0.49, 0.60, -0.23, -0.11, -0.05)^T.
$ 
We calculate the corresponding correlation matrices at \(\bd U_1\), \(\bd U_{50}\), and \(\bd U_{100}\), respectively, and focus on visualizing the heatmaps of the dynamic correlation matrices for the first 50 stocks with the largest market capitalizations. { These heatmaps are presented in Figure S.1 in Supplementary Materials S.6.  
Notable changes are evident across Figures S.1(a), S.1(b), and S.1(c), which correspond to \(\bd U_1\), \(\bd U_{50}\), and \(\bd U_{100}\), respectively.  Furthermore, a clear distinction is observed between the dynamic correlation matrices in Figures S.1(a)-S.1(c) and the static correlation matrix shown in Figure S.1(d). } These results highlight the dynamic nature of the correlations and the effectiveness of the proposed estimation approach in capturing this variability.

\subsection{The comparisons of  minimum-variance portfolios}


In our strategy, the minimum-variance portfolio is updated daily, allowing us to dynamically adjust the portfolio based on the latest market conditions. Specifically, on each trading day during the out-of-sample period, the dynamic covariance matrix is estimated using data from the most recent 100 trading days (approximately six months) as of the previous trading day. This estimated covariance matrix is then used to construct the minimum-variance portfolio as defined in (\ref{Ri}). 
After obtaining the daily out-of-sample returns of the global minimum-variance portfolio, we compute its annualized performance metrics, including the annualized average return (AVR), annualized standard deviation (STD), and annualized information ratio (IR), as defined by \cite{engle2017large}:
\[
\text{AVR} = \bar{R} \times 252, \quad 
\text{STD} = \sigma_R \times \sqrt{252}, \quad 
\text{IR} = \frac{\text{AVR}}{\text{STD}},
\]
where \(\bar{R}\) is the mean of the daily returns defined in (\ref{Ri}) during the out-of-sample period, and \(\sigma_R\) is the standard deviation of the daily returns over the same period.


\begin{table}[t]
	\centering
	\caption{Out-of-sample performance of the covariance matrix estimation.}
	\label{table-8}
	\resizebox{110mm}{!}{
    \begin{tabular}{cccc}
		\toprule
		Method & AVR(\%)    & STD(\%) & IR \\
		\midrule
		Static & -351.97  &788.45  & -0.45  \\
		FDCMs    & -731.43  & 1681.68  & -0.43  \\
        Modified FDCMs & 6.13 & \bf 6.15 & 1.00 \\
        Kernel (Mkt-RF) & -450.66 & 1860.08 & -0.24 \\
        Modified Kernel (Mkt-RF) & 5.69 & 14.03 & 0.41 \\
		\bottomrule
	\end{tabular}}
\end{table}%

We consider the following methods with soft thresholding: the static sample covariance matrix, FDCMs, modified FDCMs, the kernel-based DCM method, and its modified version as proposed in \cite{chen2016dynamic}. The market risk factor (Mkt-RF), utilized in the single-factor capital asset pricing model \citep{rossi2016capital}, represents the excess return earned for taking on additional market risk. It captures the difference between the returns of the overall market and the risk-free rate. We use Mkt-RF as the covariate for the kernel-based DCM method and its modified version. The results are summarized in Table \ref{table-8}. 
In the context of constructing global minimum-variance portfolios, \citet{engle2019large} emphasizes that the primary metric for evaluating the performance of covariance matrix estimation methods is the annualized standard deviation (STD). While high values of annualized average return (AVR) and annualized information ratio (IR) are desirable, they are considered secondary to the primary goal of minimizing STD. 
From this perspective, the modified FDCMs for dynamic covariance matrix estimation outperform the other methods. Notably, the minimum-variance portfolios constructed using the static sample covariance matrix, FDCMs, and kernel-based DCM exhibited significantly negative AVRs and high STDs. This behavior may arise from the fact that some covariance matrices estimated by these methods are not positive definite, underscoring the importance of positive definite correction in practical applications. 
Furthermore, the performance of the modified FDCMs surpasses that of the modified kernel-based DCM, suggesting that multiple factors are likely influencing the covariance matrices. This highlights the necessity of incorporating multiple covariates to effectively model the dynamic covariance matrices for this dataset.



\section{Conclusion}\label{Conclusion}

This paper introduces a novel method for estimating large dynamic covariance matrices in high-dimensional settings using honest forests. Unlike kernel-based methods, which rely on distances between covariates and the target \(\bd u\) and handle only a single covariate, our approach uses random forests to assign weights in a data-driven, nonparametric manner, accommodating multiple covariates. 
To enhance estimation, we impose sparsity through a generalized thresholding operator, penalizing the entries of the random forest-based estimators. We also establish theoretical guarantees, including uniform convergence rates for the estimator and its inverse, as well as uniform consistency in identifying sparse patterns—marking the first integration of random forests with covariance estimation in high dimensions. 
Simulations and a real-world stock market application demonstrate that our method consistently outperforms traditional static and kernel-based approaches, offering superior accuracy and robustness in finite sample settings.

\bigskip






\bibliographystyle{agsm}

\bibliography{Bibliography-MM-MC}

\clearpage

\renewcommand{\thefigure}{A\arabic{figure}} 
\renewcommand{\thetable}{A\arabic{table}}   
\renewcommand{\thealgorithm}{A\arabic{algorithm}} 
\setcounter{figure}{0} 
\setcounter{table}{0}  
\setcounter{algorithm}{0} 

\end{document}